\documentclass{style}
\usepackage{graphicx}
\usepackage{wrapfig}
\usepackage{booktabs}
\usepackage{multirow}
\usepackage{subcaption}
\usepackage{makecell}
\usepackage{url}
\usepackage{siunitx}
\usepackage{ulem}
\usepackage{enumitem}
\usepackage{quoting}
\usepackage{amsfonts}
\usepackage{float}
\usepackage{stfloats}
\usepackage{tabularx}
\usepackage{needspace}
\usepackage{pifont}

\definecolor{highlightcolor}{gray}{0.95}
\newcommand{\cmark}{\ding{51}} 
\newcommand{\xmark}{\ding{55}} 
\newcommand{\pmark}{\ding{108}} 

\tcbuselibrary{breakable}
\usepackage[svgnames]{xcolor}
\usepackage{textcase} 
\tcbset{
  prompt/.style={
    colback=OldLace,
    colframe=YellowGreen,
    width=\linewidth,
    arc=2mm,
    auto outer arc,
    breakable,
    lefttitle=0pt,      
    boxsep=1.5mm,       
    left=1.5mm, right=1.5mm, top=1.5mm, bottom=1.5mm,
    fonttitle=\sffamily,  
    fontupper=\ttfamily,     
    title={\MakeTextUppercase{#1}},
  }
}

\usepackage[utf8]{inputenc}
\usepackage{longtable} 
\usepackage{multirow}  
\usepackage{array}     
\usepackage[table]{xcolor} 
\usepackage{pifont}
\setlist[itemize]{leftmargin=*}
\definecolor{lightgray}{gray}{0.92}
\newcolumntype{C}[1]{>{\centering\arraybackslash}m{#1}}

\usepackage{times}
\usepackage{latexsym}
\usepackage{indentfirst}
\usepackage[T1]{fontenc}

\usepackage[utf8]{inputenc}

\usepackage{microtype}

\usepackage{inconsolata}
\usepackage{marvosym} 

\usepackage{graphicx}

\definecolor{ourcolor}{RGB}{240, 82, 156}
\definecolor{tablegray}{RGB}{223, 242, 252}
\definecolor{tablegreen}{RGB}{15, 203, 150}
\definecolor{tableyellow}{RGB}{250, 242, 233}
\definecolor{tableblue}{RGB}{240, 82, 156}
\definecolor{darkpink}{RGB}{139, 14, 98}

\usepackage[table,xcdraw]{xcolor}

\title{
    \includegraphics[width=0.8cm, keepaspectratio]{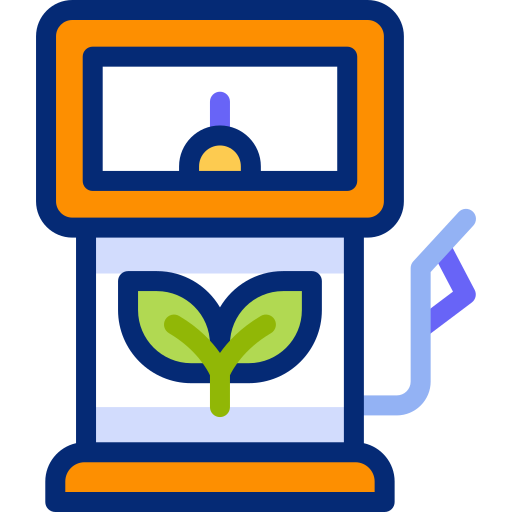} 
    \fontsize{15.5pt}{23pt}\sffamily\bfseries\color{black}
    {\textcolor{ourcolor}{G}}eneration as {\textcolor{ourcolor}{A}}uxiliary {\textcolor{ourcolor}{S}}upervision \\
    Enhancing Visual Understanding at Zero Inference Overhead via \\ 
    Decoupled Embedding Prediction
}
\newcommand{\corrmark}{\text{\Letter}}

\author{
    \fontsize{12.5pt}{19pt}\selectfont\sffamily\color{black}
    Zhongbin Guo\thanks{Equal contribution.}\thanks{Work done during an internship at ByteDance.} \quad
    Jiahao Xie\footnotemark[1] \quad
    Dongling Xiao \quad
    Qianle Wang \\
    Ruiqi Lu \quad Xiaomin He\footnotemark[2] \quad Wanxuan Sun$^\corrmark$ \quad Cheng Yang$^\corrmark$ \\[0.5em]
    \vspace{1em}
    \normalsize\sffamily\color{black} ByteDance \\
}

\email{guozhongbin66@gmail.com}
\date{\today}

\begin{document}

\maketitle

{\renewcommand{\thefootnote}{\corrmark}%
 \footnotetext{Corresponding author.}}

\thispagestyle{firstpage}

\begin{abstract}
While Multimodal Large Language Models (MLLMs) have achieved remarkable progress, visual understanding and generation are typically treated as divergent objectives. Existing unified frameworks often rely on discrete visual tokenization or diffusion objectives whose generative targets differ from the continuous representations consumed by visual understanding models, making direct transfer to enhance existing pretrained MLLMs non-trivial. 
In this work, we present \textbf{GAS}, a generation-guided training framework that reinterprets visual generation as auxiliary supervision for representation learning. 
Concretely, GAS adapts \textbf{Next Embedding Prediction (NEP)} as a cross-modal generation paradigm within a decoupled Mixture-of-Transformers (MoT) architecture. 
By maintaining a shared lower trunk and parallel upper layers, GAS lets generation losses enrich the shared visual pathway with finer spatial precision and stronger visual retention while shielding the upper understanding layers from direct generation gradients. 
To maximize this synergy, we further construct highly correlated generation tasks that demand deep cognitive grounding rather than generic synthesis alone. Across model scales and training stages, GAS improves aggregate multimodal understanding, with its most reliable gains on perception and spatial comprehension. 
Crucially, because the auxiliary generation branch is discarded after training, these gains incur \textbf{zero inference overhead}.
Extensive controlled comparisons and representation-level analyses further clarify when and why generation-guided training benefits understanding, and demonstrate the feasibility of generation-guided training as a practical route to stronger multimodal understanding.
\end{abstract}

\section{Introduction}
\label{sec:introduction}

Multimodal large language models (MLLMs) have achieved remarkable progress in visual understanding tasks, including image captioning, visual question answering, and visual grounding~\cite{li2024llava,baiQwen3VLTechnicalReport2025,zhu2025internvl3}. The prevailing paradigm encodes visual inputs through a vision encoder and projects the resulting representations into token space of large language model (LLM), which then performs reasoning and generates textual responses in an autoregressive manner \cite{liuVisualInstructionTuning2023,radford2021learning}. 
Although this architecture has proven remarkably effective for semantic-level comprehension, its training objective -- next-token prediction~(NTP) over text -- provides only indirect supervision for visual perception. The model is trained to \emph{talk about} images, yet never to \emph{learn from} them because its text objective does not directly supervise pixel- or region-level visual structure.

This text-only supervision introduces limitations from two complementary perspectives. 
On one hand, natural language inherently suffers from an \textbf{expressiveness bottleneck}: it cannot precisely encode fine-grained visual details such as exact spatial relationships, pixel-level object boundaries, physical interaction dynamics, or subtle appearance variations \cite{shao2024visual,shiri2024empirical}. When the sole training signal is text prediction, the model's visual representations are shaped only by what language can articulate, leaving the richer geometric, structural, and relational information latent and underutilized. 
On the other hand, there exist abundant \textbf{visually intensive tasks} like segmentation and image editing. Existing VLM architectures are structurally unable to consume such data for training, thereby forfeiting a vast reservoir of fine-grained visual supervision that could otherwise cultivate deeper perception and complex visual understanding capabilities. 
Furthermore, empirical evidence reveals that visual information progressively attenuates in deeper LLM layers during reasoning \cite{zhang2024visually,ganz2024question}, precisely because the text-centric objective provides no mechanism to encourage persistent retention of visual features throughout the inference chain.

A natural hypothesis emerges: if a model is additionally trained to predict visual content, such as pixel-level segmentation masks, edited images, or visual reasoning traces, then its internal representations may be encouraged to retain richer and more precise visual information that can benefit understanding. 
This intuition has motivated the recent wave of unified multimodal models (UMMs) that jointly train understanding and generation \cite{team2024chameleon,xie2025showo,zhou2025transfusion,deng2025emerging}. However, directly adopting the UMM paradigm to enhance understanding exposes two unresolved tensions:

First, current UMMs retain generation-side parameters at inference time~\cite{deng2025emerging,shen2025mammothmoda2}, incurring latency and memory costs that hinder deployment. 
Moreover, generation objectives in these systems are typically optimized for synthesis quality rather than for maximizing benefit to comprehension~\cite{fu2026lance}. Understanding performance can therefore plateau or even degrade relative to MLLMs of comparable scale, suggesting that simple joint training does not reliably transfer generative learning into understanding gains.

\begin{figure*}[t]
  \centering
  \includegraphics[width=\textwidth]{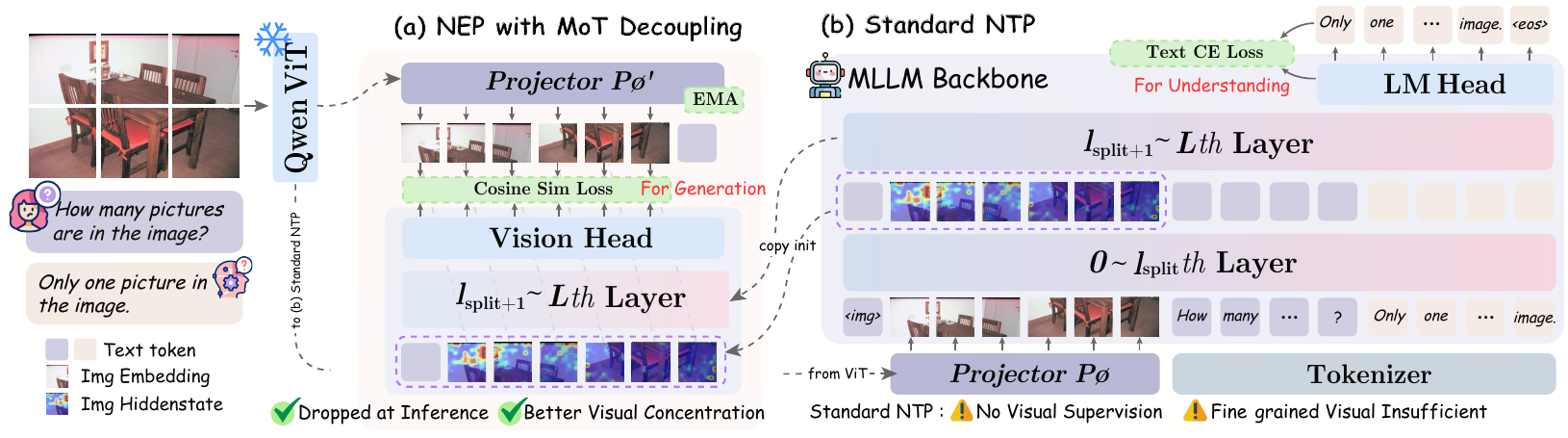}
  \caption{\textbf{Overview of our generation-guided training framework.}
    Given multimodal inputs, the understanding branch processes visual tokens from shared ViT encoder and predicts text via standard NTP paradigm. The generation branch, operating on independent MoT transformer layers, receives intermediate hidden states from the understanding backbone and predicts target image embeddings via Next Embedding Prediction (NEP). At inference time, the generation branch is entirely discarded.}
    \label{fig:framework}
\end{figure*}

Second, existing work typically reports aggregate benchmark numbers after introducing generation data, without dissecting the underlying mechanisms. Critical questions remain open: Do generation and understanding objectives inherently conflict? Which intermediate representations are most amenable to generative supervision for ``feeding back'' into understanding performance? Which \emph{types} of generation tasks enhance which facets of comprehension? And through what pathways does generation training confer its advantages?

In this work, we propose a generation-guided training framework that resolves both tensions. Our core idea is to leverage generation tasks purely as \emph{training-time auxiliary supervision} for understanding, rather than as an end goal. 
Architecturally, as shown in Figure~\ref{fig:framework}, we adopt a Mixture-of-Transformers (MoT) design that introduces a dedicated generation branch with independent transformer parameters alongside the understanding branch, so that generation gradients refine shared visual representations without directly perturbing understanding-side parameters. 
The generation branch employs the Next Embedding Prediction (NEP) paradigm to predict continuous visual embeddings autoregressively in the representation space consumed by the LLM backbone, without requiring a pixel decoder or discrete visual de-tokenizer. 
At inference time, the generation branch is entirely discarded, yielding strictly zero additional cost during inference. This deployment benefit is distinct from training cost: GAS uses approximately 11.6\% more GPU-hours than the corresponding baseline.

On the data side, we systematically construct about 10M generation samples across five primary categories and 15 subtasks, covering a spectrum from pixel-level perception to high-level reasoning, and develop an automated synthesis pipeline that produces high-quality training data without manual annotation. This framework allows us to conduct systematic analyses to elucidate \emph{how} generation enhances complex visual understanding which leads to several intriguing findings on generation-understanding conflict, task-specific ablation and representation-level diagnostics.

We finally present the \textbf{GAS} model family, validating the framework on 2B- and 4B-parameter backbones. GAS improves aggregate visual understanding at both scales, while the per-benchmark transfer is task-dependent. Our main contributions are as follows:
\begin{itemize}[leftmargin=1.5em,itemsep=2pt]
    \item We combine continuous NEP supervision with a removable MoT generation branch, transferring generation-side learning to the deployed understanding model at zero inference overhead.
    \item We construct diverse generation task categories with an automated data synthesis pipeline, and show that tasks with stronger latent correlations to understanding deliver the largest gains, while combining them yields further, complementary improvements on perception- and reasoning-oriented benchmarks.
    \item We provide systematic analyses of parameter isolation, layer-wise supervision injection, per-task contribution, representation-level diagnostics, matched-budget controls and repeated-run statistics, clarifying when and how generation enhances understanding.
\end{itemize}

\section{Methodology}
\label{sec:method}

Our framework is designed around a single guiding principle: generation tasks should serve as \emph{training-time auxiliary supervision} that enriches the model's visual representations, without altering the inference-time architecture or cost. To realize this, we introduce three tightly integrated components: (1)~Next Embedding Prediction (NEP) paradigm that formulates image generation as autoregressive embedding prediction within the same representation space used for understanding; (2)~Mixture-of-Transformers (MoT) architecture that decouples generation and understanding parameters while preserving a shared visual representation pathway; and (3)~systematic multi-type generation data construction strategy that maximizes the complementary benefits of diverse visual tasks.

\subsection{Next Embedding Prediction (NEP)}
\label{sec:nep}

Existing unified models typically adopt either discrete visual tokenization~\cite{wangMultimodalLearningNexttoken2026,wu2025janus} or diffusion-based generation~\cite{yang2026mmada,xie2025showo}. 
Both approaches introduce representation spaces fundamentally disjoint from the continuous embeddings consumed by the understanding branch, severely limiting how much generation training can benefit the shared visual representations~\cite{team2026longcat}.

To bridge this gap, we propose \textbf{Next Embedding Prediction (NEP)} as a cross-modal generation paradigm. Recent literature~\cite{xu2025next} has demonstrated that predicting continuous embeddings autoregressively serves as an exceptionally strong visual learner. Building upon this insight, our core motivation is to formulate image generation as autoregressive embedding prediction within the \emph{exact same continuous representation space} used by the LLM backbone. 

This formulation anchors the generative target directly in the continuous input space of the language model, and preserves a unified representation manifold for both perception and generation. 
Unlike discrete tokenization which maps images into a separate categorical codebook, or diffusion objectives defined in a specialized synthesis-oriented latent space, NEP avoids a separate target-to-LLM translation interface. 
The design is therefore intended to make auxiliary visual supervision more directly compatible with the representations used for downstream understanding; Section~\ref{sec:analysis} tests this claim against fixed-architecture and related-objective controls.

Contemporaneous UniHetero~\cite{chen2025unihetero} likewise finds that autoregressing on LLM input embeddings is effective at large data scale. NEP is differentiated by using this representation as the target for \emph{instruction-conditioned cross-image prediction} across diverse generation tasks, and by coupling it to an asymmetric MoT branch whose upper-layer gradients are isolated from the understanding path and whose parameters are removed after training.

\noindent\textbf{Formulation and Supervision Target.} 
Unlike single-image self-supervised pretraining~\cite{xu2025next}, NEP operates in a conditional, multi-task setting. Given a multimodal context containing an input image $I$ and diverse linguistic instructions (e.g., for segmentation or grounding), let $\mathbf{x}_{\text{ctx}}$ denote the concatenated sequence of text tokens and source image embeddings. For a generative task with a target image $I^{\text{tgt}}$, the model autoregressively predicts the target embedding sequence $\mathbf{\hat{z}}^{\text{tgt}}$. 

Crucially, rather than introducing an external reconstruction decoder to predict raw pixels, we extract ground-truth embeddings using the identical vision encoder followed by a visual-language projector. To ensure the target resides in the exact LLM input space, the target embeddings are defined as $\mathbf{z}^{\text{tgt}} = \text{Projector}(\text{ViT}(I^{\text{tgt}}))$. The model predicts each embedding conditioned on the context and previously generated embeddings:
\begin{equation}
    \hat{z}^{\text{tgt}}_i = f_{\text{gen}}\big(\mathbf{x}_{\text{ctx}}, \hat{z}^{\text{tgt}}_{<i}; \Theta_{\text{gen}}\big),
\end{equation}
where $f_{\text{gen}}$ denotes the generative forward pass of the network parameterized by $\Theta_{\text{gen}}$. To optimize this objective, both the predicted and target embeddings are first $\ell_2$-normalized. The generation loss is then computed as the cosine distance in this continuous representation space:
\begin{equation}
    \mathcal{L}_{\text{gen}} = \frac{1}{N} \sum_{i=1}^{N} \left( 1 - \frac{\hat{z}^{\text{tgt}}_i \cdot z^{\text{tgt}}_i}{\|\hat{z}^{\text{tgt}}_i\|_2 \|z^{\text{tgt}}_i\|_2} \right).
\end{equation}
By optimizing directly within the understanding interface rather than pursuing photorealistic pixel synthesis, we explicitly prioritize the enrichment of deep structural representations over visual fidelity.

\noindent\textbf{Target Stabilization via EMA.} 
A critical challenge in predicting representations projected into the LLM space is that the active visual projector parameters are continuously updated during joint training~\cite{chen2025unihetero}. Utilizing a dynamic projector to extract targets can lead to supervision drift and representation collapse. To prevent this, we maintain a \textit{target projector} whose weights are updated via an Exponential Moving Average (EMA) of the active projector's weights. The stable target embeddings $\mathbf{z}^{\text{tgt}}$ are thus computed by passing the target image through the frozen ViT followed by this EMA target projector. Correspondingly, a dedicated vision-head is applied to the output hidden states of the LLM's generation branch to predict these stable targets. 
As detailed in Section~\ref{sec:mot}, the decoupled architecture with this robust NEP formulation safely injects precise spatial information into the shared representations, and further mitigates direct interference with the upper understanding layers.
\begin{figure*}[t]
  \centering
  \includegraphics[width=\textwidth]{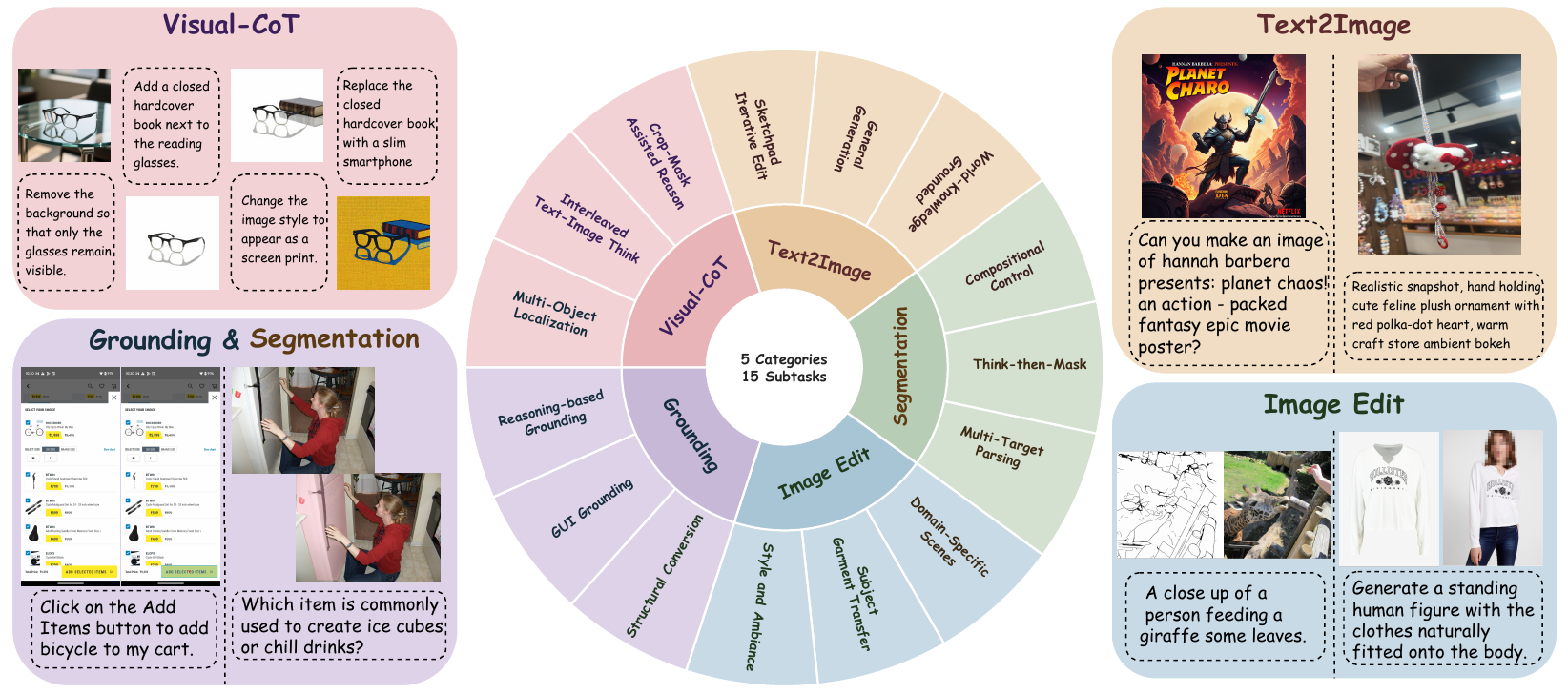}
  \caption{\textbf{Overview of generation training data.}
    Five primary task categories spanning 15 subtasks, jointly covering the fine-grained visual capabilities targeted by our generation-guided training.}
    \label{fig:data_overview}
\end{figure*}

\subsection{MoT Decoupled Architecture}
\label{sec:mot}

A straightforward approach to introducing generation supervision is to add the NEP loss directly onto the shared LLM backbone. However, our experiments (Section~\ref{sec:analysis}) and prior works reveal that this strategy degrades understanding capability~\cite{deng2025emerging,li2025unifork,hao2025uni}: the generative and understanding objectives impose conflicting optimization pressures, ultimately diluting the abstract semantic representations necessary for complex reasoning. To resolve this tension, we adopt the Mixture-of-Transformers (MoT) architecture~\cite{liang2025mixtureoftransformers}, which decouples the upper task-specific parameters while preserving a shared lower trunk for indirect knowledge transfer.

\noindent\textbf{Decoupled Architecture.} To preserve visual structures that are typically abstracted in deeper layers focused on semantics, the model topology bifurcates at an intermediate layer $l_{\text{split}}$. 
The understanding branch continues to process the sequence through the remaining upper layers ($l > l_{\text{split}}$), optimized solely by the text cross-entropy loss ($\mathcal{L}_{\text{und}}$). Concurrently, we extend the backbone with a parallel generation branch comprising $N_{\text{gen}}$ transformer layers. To guarantee a reasonable initialization and stabilize early training, the parameters of these generation layers ($\Theta_{\text{gen}}$) are directly copied from the corresponding pre-trained layers of the understanding backbone. This branch extracts hidden states from layer $l_{\text{split}}$ to autoregressively produce NEP predictions:
\begin{equation}
    \mathbf{H}_{\text{gen}} = \text{Transformer}_{\text{gen}}(\mathbf{H}^{(l_{\text{split}})}; \Theta_{\text{gen}})
\end{equation}
\begin{equation}
    \hat{z}^{\text{tgt}}_i = \text{VisionHead}(\mathbf{h}_{\text{gen}, i}; \Theta_{\text{head}}),
\end{equation}
where $\mathbf{H}^{(l_{\text{split}})}$ denotes the sequence of hidden states at layer $l_{\text{split}}$, and $\mathbf{h}_{\text{gen}, i}$ is the $i$-th token representation in the generation output $\mathbf{H}_{\text{gen}}$. The generation branch is parameterized by $\Theta_{\text{gen}}$ and $\Theta_{\text{head}}$, supervised entirely by the NEP loss ($\mathcal{L}_{\text{gen}}$). 

\noindent\textbf{Mechanism of Synergy.} A critical question arises: how can generative training enhance understanding if their parameters diverge? The key lies in the asymmetric gradient flow across the shared first $l_{\text{split}}$ layers. This partial decoupling induces two powerful, complementary effects:

\noindent\textbf{(1) Direct Refinement of Shared Representations.} During joint training, gradients from $\mathcal{L}_{\text{gen}}$ backpropagate through the generation branch into the visual projection and the shared trunk ($l \le l_{\text{split}}$). This forces these lower layers to produce intermediate states $\mathbf{H}^{(l_{\text{split}})}$ that are enriched with precise, fine-grained visual details. Because $\mathbf{H}^{(l_{\text{split}})}$ acts as the direct input to the upper understanding branch, this visual enrichment is seamlessly inherited. 

\noindent\textbf{(2) Implicit Adaptation of Upper Layers.} While the upper understanding layers ($l > l_{\text{split}}$) remain shielded from generative gradients, they are continuously fed the visually enriched representations from the shared trunk. Consequently, they are compelled to adapt under $\mathcal{L}_{\text{und}}$ to fully exploit these high-quality features. Empirically (Section~\ref{sec:analysis}), this adaptation manifests as sharpened attention heads that reliably anchor onto question-relevant visual regions, thereby enhancing evidence retrieval during multi-step reasoning.

\noindent\textbf{Training and Inference.} The overall objective minimizes the weighted sum $\mathcal{L} = \mathcal{L}_{\text{und}} + \lambda \cdot \mathcal{L}_{\text{gen}}$. Crucially, at inference time, the entire generation branch ($\Theta_{\text{gen}}$ and $\Theta_{\text{head}}$) is discarded. Only the understanding branch performs the forward pass, rendering the inference cost, latency, and memory footprint strictly identical to the standard baseline model.

\subsection{Generation Task Construction}
\label{sec:task_construction}

Prior unified multimodal works typically introduce generation data in the form of generic text-to-image synthesis or simple editing, treating generation merely as a self-contained skill. Under such configurations, understanding performance gains are often negligible or even negative. We argue that generic synthesis tasks, such as ``generate an image of a cat'', provide little training signal relevant to complex visual reasoning, as they merely demand a superficial alignment between basic nouns and visual concepts. 

The core insight of our framework is that \textbf{generation can better enhance understanding when the generative act inherently depends on deep multimodal comprehension.} Rather than treating generation as a straightforward rendering process, we purposefully formulate it as the terminal output of a complex cognitive task. If a model must execute precise spatial localization, compositional scene analysis, or multi-step logical deduction just to figure out \emph{what} or \emph{where} to generate, the generation objective acts as a powerful forcing function. It compulsorily drives the network to develop the exact fine-grained perceptual and reasoning capabilities that pure text supervision fails to provide.

\begin{figure*}[t]
  \centering
  \includegraphics[width=\textwidth]{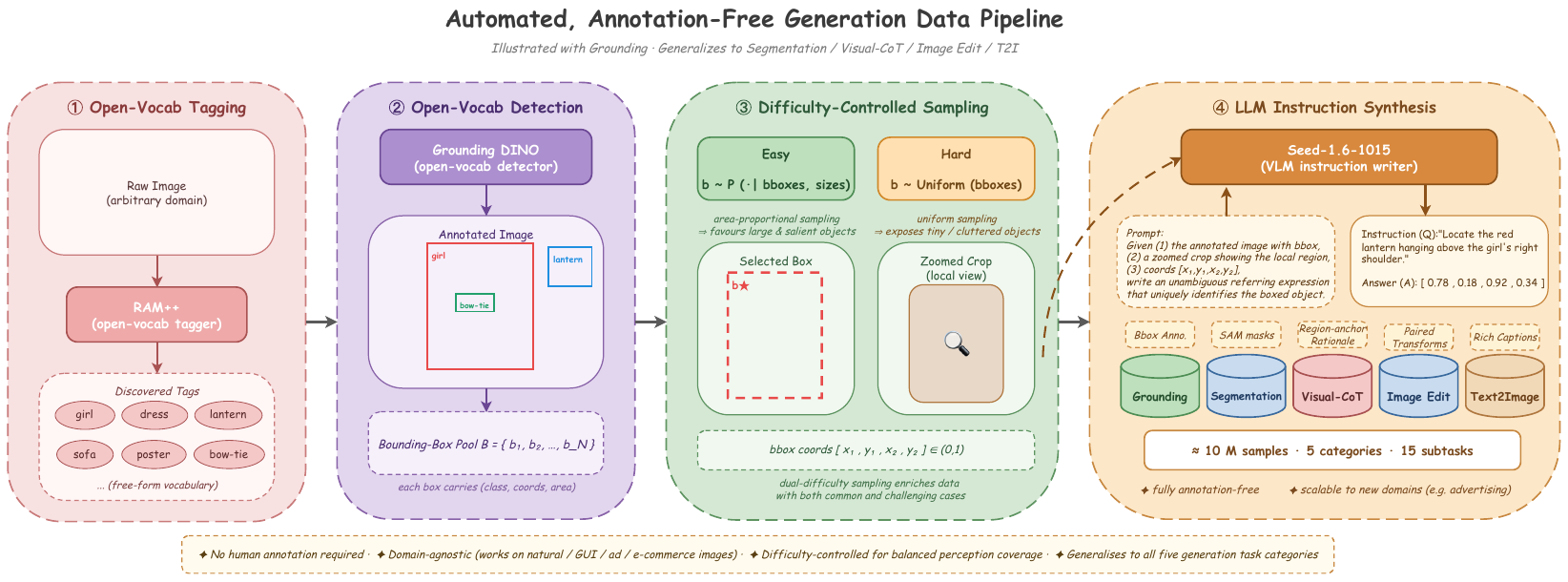}
  \caption{\textbf{Generation task construction pipeline.}
    Illustration of our automated, annotation-free data construction pipeline, taking grounding as a representative example.}
    \label{fig:data_pipeline}
\end{figure*}

Guided by this principle, we select candidate generation tasks based on their \emph{latent cognitive dependency} with the target understanding capabilities. We construct five generation tasks with 15 subtasks in total (illustrated in Figure~\ref{fig:data_overview}):

\begin{itemize}[leftmargin=0.8em,itemsep=3pt]
    \item \textbf{Grounding}: Region localization tasks that compel the model to perfectly resolve complex linguistic referring expressions before predicting absolute spatial coordinates, directly cultivating fine-grained spatial reasoning.
    \item \textbf{Segmentation}: Pixel-level mask prediction tasks that force the model to comprehend semantic boundaries and physical extents based on intricate textual instructions, reinforcing instance-level discrimination.
    \item \textbf{Image Editing}: Conditional image transformation tasks (e.g., style transfer, object removal) that require a deep, compositional understanding of scene layouts and object interdependencies before executing the edit.
    \item \textbf{Visual Chain-of-Thought}: Multi-step reasoning tasks that demand the model to actively generate intermediate visual evidence (e.g., highlighting key regions) to anchor multi-step logical deduction, simulating a ``thinking with images'' process.
    \item \textbf{Text-to-Image (T2I)}: Carefully curated image synthesis tasks grounded in rich world knowledge and dense structural prompts, establishing foundational cross-modal alignment beyond naive concept mapping.
\end{itemize}

Crucially, our ablation studies (Section~\ref{sec:analysis}) confirm that these task categories provide \emph{complementary supervision signals}. Empirically, on Counting \& Spatial benchmarks the gain from combining all categories tends to exceed the best single-task gain, suggesting that diverse, comprehension-correlated generation tasks are not redundant; rather, they jointly cover different facets of fine-grained visual structure that no single task fully captures. 

Building on this finding, we develop a scalable data synthesis framework to produce high-quality generation data across diverse domains without manual annotation. Taking grounding as a representative example, our automated pipeline operates in four stages: (1)~open-vocabulary tagging via RAM++ to discover all salient objects; (2)~open-vocabulary detection via Grounding DINO to localize each tag; (3)~difficulty-controlled sampling (area-proportional for easy, uniform for hard) to select target boxes; and (4)~LLM-based instruction synthesis using the annotated image and a zoomed crop to generate unambiguous referring expressions. This pipeline generalizes to arbitrary image domains and enables scaling of generation data to new scenarios (e.g., advertising) with minimal engineering effort. 

Combining the above task selection principle with this automated construction and filtering pipeline, we ultimately scaled the total generation training corpus to approximately 10M samples spanning all five task categories, and mixed them with understanding data for training.

\section{Experiments}
\label{sec:experiments}

\newcounter{qualbox}
\renewcommand{\thequalbox}{\arabic{qualbox}}
\newcommand{\qualboxref}[1]{Box~\ref{#1}}

\tcbset{
  qualbox/.style={
    breakable,
    enforce breakable,
    fontupper=\small,
    fonttitle=\small,
    before skip=4pt,
    after skip=4pt,
    boxsep=2pt,
    left=4pt, right=4pt, top=2pt, bottom=2pt,
    pad at break*=2pt,
  }
}

\subsection{Experiments Setup}

\textbf{Model Configuration.} To ensure a controlled environment devoid of data contamination from existing VLM weights, we build our main models from scratch. 
Adopting the Qwen3-VL architecture~\cite{baiQwen3VLTechnicalReport2025} for the understanding branch, we utilize Qwen3-VL-ViT as a visual encoder initialized with pretrained weights, which remains frozen throughout all training stages. For the LLM backbone, we employ the 2B and 4B Qwen3~\cite{yang2025qwen3} models. 
Note that we use the raw LLM weights without any prior multimodal alignment. Additionally, the ViT-LLM projector is randomly initialized, ensuring all visual-language capabilities are acquired solely through our training pipeline.
The MoT generation branch consists of $L_{\text{mot}}$ transformer layers initialized from the corresponding pre-trained layers of the understanding backbone, with hidden states extracted from the intermediate layer $l_s \approx L/2 = 14$ of the understanding backbone (where $L$ is the total layer count).

\textbf{Implementation Details.}
\label{sec:implementation}
Training proceeds in two stages. In \textbf{Stage~1 (MoT Align)}, we freeze all understanding-side parameters and train only the MoT generation branch using T2I generation data so as to activate the generation pathway and align it with the existing visual representation space without perturbing the pretrained understanding capability. In \textbf{Stage~2 (Joint Training)}, both understanding and generation data are used simultaneously. The understanding branch is optimized with text cross-entropy loss, while the generation branch is optimized with NEP loss under a progressive weighting schedule that linearly ramps $\lambda$ from an initial value of 0.015 to the target weight 1.0 over 4k steps. 

This setting allows generation gradients to refine the shared trunk while leaving the ViT frozen, thereby enabling the indirect transfer mechanism described in Section~\ref{sec:mot}.
On our HPC Cluster, the final baseline uses 2{,}208 GPU-hours, whereas GAS uses 2{,}464 GPU-hours ($+11.6\%$); the corresponding ablation runs use 544 and 608 GPU-hours ($+11.8\%$). These totals include the generation branch, vision head, and EMA target projector. All three auxiliary components are removed at inference.

As for hyper-parameters, we use the AdamW~\cite{kingma2014adam} optimizer with a cosine learning rate schedule, a peak learning rate of $5\times10^{-5}$ for both understanding branch and the MoT generation branch. We maintain a target projector EMA with a decay rate of 0.999 for training stability, following the update rule $\theta_{\text{EMA}}^{(t)} = 0.999 \cdot \theta_{\text{EMA}}^{(t-1)} + 0.001 \cdot \theta^{(t)}$. Training is conducted on 32 GPUs with DeepSpeed ZeRO-2~\cite{rajbhandari2020zero}, using a global batch size of 128. The maximum sequence length is 8192 tokens.


\subsection{Evaluation Setup}

To comprehensively assess the impact of generation-guided training on complex visual understanding, we organize our evaluation into four capability dimensions, each targeting aspects where fine-grained visual perception and deep reasoning are essential:

\textbf{(1) General Perception (Multi-Image).}
We evaluate holistic multimodal understanding and multi-image reasoning on {MME}~\cite{fu2026mme}, {MMMU}~\cite{yue2024mmmu}, {BLINK}~\cite{fu2024blink} and {RealWorldQA}. These benchmarks assess broad visual perception, world knowledge integration, and multi-image comprehension, capabilities where richer visual representations should yield direct benefits.

\textbf{(2) Visual Reasoning.}
We evaluate mathematical and logical visual reasoning on {CharXiv}~\cite{wang2024charxiv}, {DynaMath}~\cite{zou2025dynamath}, {MathVision}~\cite{wang2024measuring}, {MathVista}~\cite{lu2024mathvista}, {LogicVista}~\cite{xiao2024logicvista} and {VisuLogic}~\cite{xu2026visulogic}. These benchmarks require precise visual parsing of charts, diagrams, and geometric figures combined with multi-step logical inference, precisely the type of complex reasoning that should benefit from generation-enhanced visual representations.

\textbf{(3) Counting \& Spatial Understanding.}
We evaluate fine-grained spatial perception on {CountBench}~\cite{paiss2023countclip,beyer2024paligemma} and {CV-Bench}~\cite{tong2024cambrian}, which specifically probe object counting accuracy and spatial relationship understanding, capabilities directly trained by our grounding and segmentation generation tasks.

\textbf{(4) Video Understanding.}
We evaluate temporal reasoning on {Video-MME}~\cite{fu2025video} and {MVBench}~\cite{li2024mvbench}. Video understanding requires sustained visual attention across frames and temporal reasoning, testing whether the improved visual information retention induced by generation training generalizes to the temporal domain.

\subsection{Main Results}
\label{sec:main_results}

\newcommand{\mycell}[2]{%
  \rotatebox{90}{%
    \parbox{2.0cm}{%
      \setlength{\baselineskip}{0.5em}%
      \textbf{\scriptsize{#1}}\\
      \footnotesize{\textcolor{gray}{#2}}%
    }%
  }%
}

\newcommand{\newcell}[1]{%
  \rotatebox{90}{%
    \parbox{2.0cm}{%
      \setlength{\baselineskip}{0.5em}%
      \textbf{\scriptsize{#1}}
    }%
  }%
}

\begin{table*}[!ht]
    \renewcommand{\arraystretch}{0.98}
    \centering
    \setlength{\tabcolsep}{2.5pt}
    \resizebox{\textwidth}{!}{
    \begin{tabular}{@{}lc
    >{\columncolor{tableyellow!50}}c
    >{\columncolor{tableyellow!50}}c
    >{\columncolor{tableyellow!50}}c
    >{\columncolor{tableyellow!50}}c
    >{\columncolor{tablegreen!10}}c
    >{\columncolor{tablegreen!10}}c
    >{\columncolor{tablegreen!10}}c
    >{\columncolor{tablegreen!10}}c
    >{\columncolor{tablegreen!10}}c
    >{\columncolor{tablegreen!10}}c
    >{\columncolor{tablegreen!10}}c
    >{\columncolor{tableblue!10}}c
    >{\columncolor{tableblue!10}}c
    >{\columncolor{tableblue!10}}c
    >{\columncolor{tablegray}}c
    >{\columncolor{tablegray}}c@{}}
        \textbf{Model} & \textbf{\#Params}
        & \mycell{MME}{test~\cite{fu2026mme}}
        & \mycell{MMMU}{val~\cite{yue2024mmmu}}
        & \mycell{BLINK}{test~\cite{fu2024blink}}
        & \mycell{RealWorldQA}{test}
        & \mycell{CharXiv}{DS~\cite{wang2024charxiv}}
        & \mycell{CharXiv}{RS~\cite{wang2024charxiv}}
        & \mycell{DynaMath}{test~\cite{zou2025dynamath}}
        & \mycell{MathVision}{mini~\cite{wang2024measuring}}
        & \mycell{MathVista}{mini~\cite{lu2024mathvista}}
        & \mycell{LogicVista}{mini~\cite{xiao2024logicvista}}
        & \mycell{VisuLogic}{test~\cite{xu2025visulogic}}
        & \mycell{CountBenchQA}{test~\cite{paiss2023countclip,beyer2024paligemma}}
        & \mycell{CV-Bench}{2D~\cite{tong2024cambrian}}
        & \mycell{CV-Bench}{3D~\cite{tong2024cambrian}}
        & \mycell{Video-MME}{test~\cite{fu2025video}}
        & \mycell{MVBench}{test~\cite{li2024mvbench}}\\
        \midrule

        \multicolumn{18}{@{}l}{\textbf{\textit{Understanding-only MLLMs}}} \\
        \textcolor{gray}{Thyme-VL$^{*}$}~\cite{zhang2026thyme}          & \textcolor{gray}{7B} & \textcolor{gray}{-} & \textcolor{gray}{-} & \textcolor{gray}{56.1} & \textcolor{gray}{70.2} & \textcolor{gray}{-} & \textcolor{gray}{-} & \textcolor{gray}{-} & \textcolor{gray}{27.6} & \textcolor{gray}{70.0} & \textcolor{gray}{-} & \textcolor{gray}{23.4} & \textcolor{gray}{-} & \textcolor{gray}{-} & \textcolor{gray}{-} & \textcolor{gray}{-} & \textcolor{gray}{-} \\
        LLaVA-v1.5$^{*}$~\cite{liu2024improved}          & 7B & - & 35.7 & - & 54.8 & - & - & 16.6 & 8.52 & - & - & 24.6 & - & - & - & - & - \\
        Qwen2.5-VL~\cite{baiQwen25VLTechnicalReport2025}                 & 3B & 2204 & 49.2 & 48.6 & 64.1 & - & - & - & 14.8 & - & - & - & - & - & - & 54.1 & 61.3 \\
        Qwen3-VL~\cite{baiQwen3VLTechnicalReport2025}                 & 2B & 2026 & 48.1 & 54.9 & 63.7 & 65.0 & 38.8 & 30.2 & 20.7 & 51.4 & 36.5 & 26.4 & 86.9 & 70.2 & 86.8 & 48.1 & 50.4 \\
        \midrule

        \multicolumn{18}{@{}l}{\textbf{\textit{Unified Models}}} \\
        \textcolor{gray}{BAGEL$^{*}$}~\cite{deng2025emerging}                                        & \textcolor{gray}{7B+7B} & \textcolor{gray}{2381} & \textcolor{gray}{55.3} & \textcolor{gray}{-} & \textcolor{gray}{72.8} & \textcolor{gray}{-} & \textcolor{gray}{-} & \textcolor{gray}{-} & \textcolor{gray}{-} & \textcolor{gray}{73.2} & \textcolor{gray}{44.3} & \textcolor{gray}{41.7} & \textcolor{gray}{82.5} & \textcolor{gray}{-} & \textcolor{gray}{-} & \textcolor{gray}{-} & \textcolor{gray}{-} \\
        \textcolor{gray}{Emu3$^{*}$}~\cite{wangMultimodalLearningNexttoken2026}    & \textcolor{gray}{8B} & \textcolor{gray}{-} & \textcolor{gray}{31.6} & \textcolor{gray}{-} & \textcolor{gray}{57.4} & \textcolor{gray}{-} & \textcolor{gray}{-} & \textcolor{gray}{-} & \textcolor{gray}{-} & \textcolor{gray}{-} & \textcolor{gray}{-} & \textcolor{gray}{24.7} & \textcolor{gray}{65.2} & \textcolor{gray}{-} & \textcolor{gray}{-} & \textcolor{gray}{-} & \textcolor{gray}{-} \\
        \textcolor{gray}{MetaMorph$^{*}$}~\cite{tong2025metamorph}                           & \textcolor{gray}{8B} &  \textcolor{gray}{-} & \textcolor{gray}{41.8} & \textcolor{gray}{-} & \textcolor{gray}{58.3} & \textcolor{gray}{-} & \textcolor{gray}{-} & \textcolor{gray}{-} & \textcolor{gray}{-} & \textcolor{gray}{-} & \textcolor{gray}{-} & \textcolor{gray}{-} & \textcolor{gray}{-} & \textcolor{gray}{-} & \textcolor{gray}{-} & \textcolor{gray}{-} & \textcolor{gray}{48.8} \\
        Janus-Pro~\cite{chen2025janus}                         & 7B & 1978$^{*}$ & 41.0$^{*}$ & 39.7 & 58.0$^{*}$ & 49.9 & 24.4 & 24.7 & 15.8 & 45.3$^{*}$ & 28.6 & 23.8 & 82.8 & 68.7 & 66.5 & - & - \\
        Show-o$^{*}$~\cite{xie2025showo}                           & 1.3B & - & 26.7 & - & - & - & - & - & - & - & - & - & - & - & - & - & - \\
        Show-o2$^{*}$~\cite{xie2026showo}                           & 1.5B & - & 37.1 & - & 56.5 & - & - & - & - & - & - & - & - & - & - & - & 50.6 \\
        Cheers$^{*}$~\cite{zhang2026cheers}                           & 1.5B & - & 36.0 & - & 60.9 & - & - & - & - & 50.5 & - & - & - & - & - & - & - \\
        \midrule

        \multicolumn{18}{@{}l}{\textbf{\textit{Ours}}} \\
        
        \textcolor{ourcolor}{GAS}                           & 2B & 2003 & 46.6 & 48.0 & 64.8 & 64.4 & 34.3 & 47.9 & 20.4 & 56.4 & 32.7 & 28.6 & 90.1 & 73.2 & 73.8 & 46.2 & 47.8 \\
        \textcolor{gray}{~~~-baseline}    & \textcolor{gray}{} & \textcolor{gray}{1986} & \textcolor{gray}{46.9} & \textcolor{gray}{46.7} & \textcolor{gray}{64.1} & \textcolor{gray}{64.0} & \textcolor{gray}{34.6} & \textcolor{gray}{46.2} & \textcolor{gray}{19.1} & \textcolor{gray}{54.4} & \textcolor{gray}{36.0} & \textcolor{gray}{26.8} & \textcolor{gray}{87.7} & \textcolor{gray}{69.9} & \textcolor{gray}{73.2} & \textcolor{gray}{45.2} & \textcolor{gray}{49.0} \\
        
        \textcolor{ourcolor}{GAS}                           & 4B & 2146 & 57.0 & 56.9 & 70.3 & 78.4 & 40.5 & 55.2 & 27.0 & 68.1 & 37.1 & 23.6 & 90.8 & 75.4 & 83.6 & 54.4 & 52.4 \\
        \textcolor{gray}{~~~-baseline}    & \textcolor{gray}{} & \textcolor{gray}{2092} & \textcolor{gray}{54.9} & \textcolor{gray}{57.8} & \textcolor{gray}{70.2} & \textcolor{gray}{77.9} & \textcolor{gray}{40.3} & \textcolor{gray}{57.6} & \textcolor{gray}{25.3} & \textcolor{gray}{65.6} & \textcolor{gray}{37.1} & \textcolor{gray}{26.1} & \textcolor{gray}{90.1} & \textcolor{gray}{75.1} & \textcolor{gray}{82.1} & \textcolor{gray}{55.0} & \textcolor{gray}{54.2} \\

    \end{tabular}
    }%
    \vspace{0.6em}
    \caption{\textbf{Overall results} for a range of understanding-only MLLMs, Unified Models, baselines and our GAS family. 
    $^{*}$ denotes metrics taken directly from public reports rather than reproduced by us.}
    \label{tab:overall_results}
    \vspace{0.4em}
\end{table*}

Table~\ref{tab:overall_results} reports the main results of GAS, its matched understanding-only baselines, public MLLM and unified-model references across all benchmarks. The paired GAS/baseline rows are the primary evidence for attribution. All improvements of GAS are achieved with zero additional inference cost since the generation branch is discarded after training.

At the same 2B scale, GAS is highly competitive with baseline and other open-source models while showing markedly stronger reasoning and counting/spatial behavior: DynaMath increases from 46.2 to \textbf{47.9} ($+1.7$pp), MathVista from 54.4 to \textbf{56.4} ($+2.0$pp), CountBenchQA from 87.7 to \textbf{90.1}, CV-Bench-2D from 69.9 to \textbf{73.2}, and VisuLogic from 26.8 to \textbf{28.6}, all directly aligned with the spatial perception and visual reasoning capabilities targeted by our grounding, segmentation, and Visual-CoT generation tasks. 
Scaling to 4B, GAS establishes the strongest overall results in the unified family, leading on most of the 16 benchmarks we report (e.g., MMMU 57.0, CharXiv-DS 78.4 / RS 40.5, CountBenchQA 90.8) and surpassing significantly larger unified models such as BAGEL (7B+7B) and Emu3 (8B). The lone exception is VisuLogic at 4B (23.6), where the larger backbone shows a benchmark-specific regression that does not transfer to other reasoning suites, suggesting an idiosyncratic sensitivity of this puzzle-style benchmark to backbone scale.
Together, these results show that our generation-guided framework consistently transfers fine-grained perception and reasoning gains to the understanding branch across both scales.

Qualitative Case~\ref{box:cvbench} and additional cases in Appendix~\ref{app:more_qualitative_cases} illustrate model behavior on CV-Bench, DynaMath, VisuLogic, and RealWorldQA. 
These selected examples show improved spatial localization, recognition of fine-grained visual details (particularly for small objects and subtle differences), and more reliable visual reasoning chains that reference image evidence. 
These qualitative patterns align well with the quantitative gains observed across our four evaluation dimensions, and suggest that the generation-derived supervision successfully transfers pixel-level and region-level perception capabilities to the understanding branch. 
In the following section, we use controlled ablations and representation-level analyses to further examine how and when generation training produces these improvements.

\vspace{1.5em}
\refstepcounter{qualbox}\label{box:cvbench}
\begin{tcolorbox}[qualbox, title = {Qualitative Case~\thequalbox: CV-Bench~\cite{tong2024cambrian}}]

\vspace{\medskipamount}
{\includegraphics[width=0.4\linewidth]{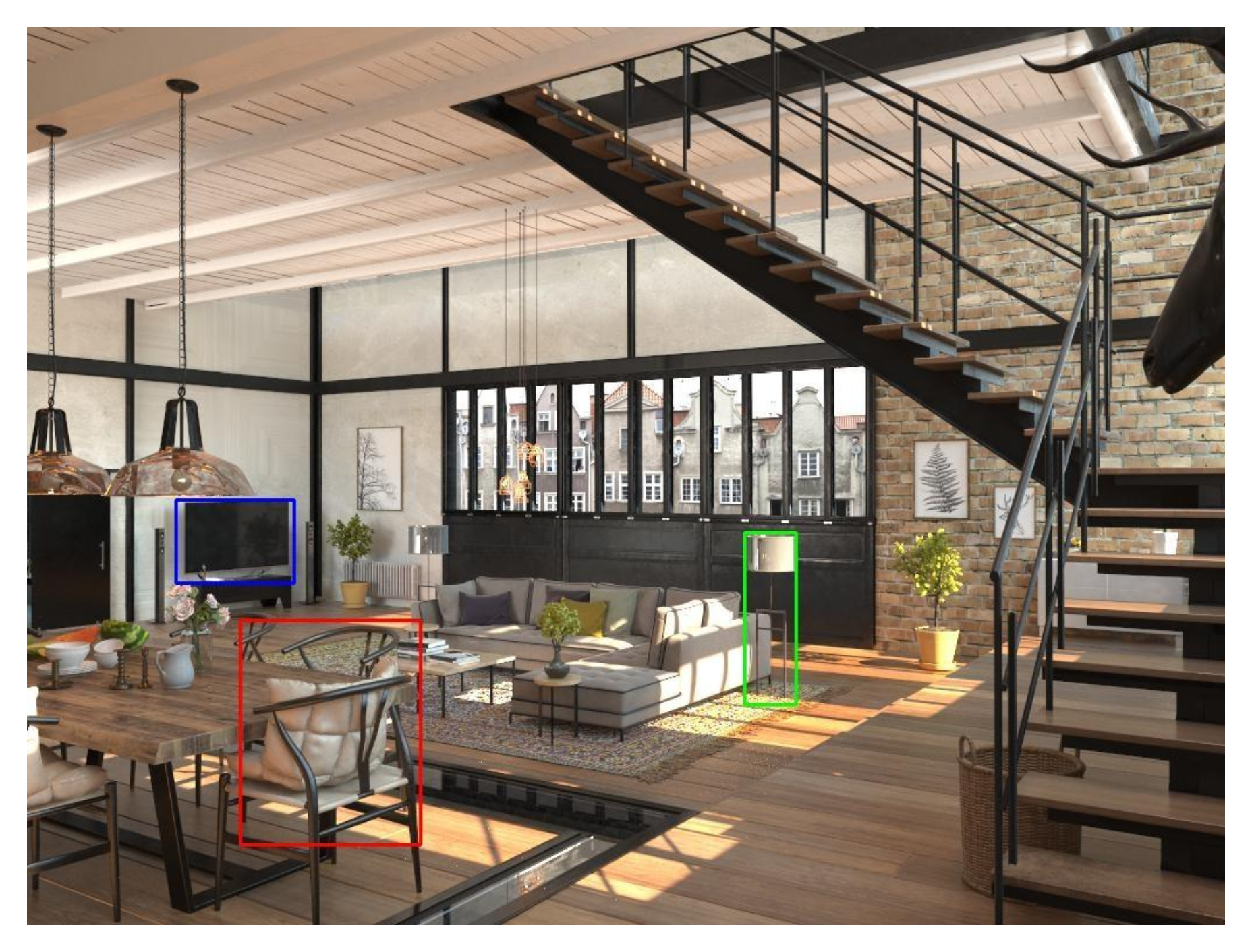}}
\vspace{\medskipamount}

\textbf{Question: }Estimate the real-world distances between objects in this image. Which object is closer to the chair (highlighted by a red box), the television (highlighted by a blue box) or the lamp (highlighted by a green box)?
 A. television, 
 B. lamp

\noindent\rule{\linewidth}{0.5pt}
\vspace{1em}
\textbf{GAS-4B: }

{\scriptsize\ttfamily
\vspace{1em}
\textbf{\textcolor{blue}{<think>}}

To determine which object is closer to the chair (highlighted by the red box), we need to analyze the spatial relationships in the image:
 
 1. **Chair (red box)**: This is located in the foreground, near the dining table.

 2. **Television (blue box)**: This is positioned on a stand against the wall, further back in the room.

 3. **Lamp (green box)**: This is placed on the floor near the sofa, closer to the center of the room but still farther back than the chair.
 
\texttt{Analysis:}
 - The **chair** is in the foreground, very close to the viewer.
 
 - The **television** is on a stand against the wall, which is significantly farther away from the chair.

 - The **lamp** is on the floor near the sofa, which is also farther away from the chair compared to the chair's position.
 
 \texttt{Conclusion:}
 - The **chair** is the closest object to itself, and the **lamp** is closer to the chair than the **television**.
 
 Thus, the correct answer is:
 
 **B. lamp**

\textbf{\textcolor{blue}{</think>}}\\

The final answer is \textbf{\textcolor{tablegreen}{$\boxed{B}$}}.
}

\end{tcolorbox}
\section{Analysis}
\label{sec:analysis}

We conduct a series of controlled experiments to dissect the mechanisms underlying generation-guided training. 
Unless otherwise noted, analyses use the 2B-parameter model with 2.5B understanding and 2.5B generation tokens and the same evaluation suite from Section~\ref{sec:experiments}. 
We organize the investigation around five core questions: (1) Does GAS provide consistent benefits, transfer across training stages and backbone capability levels? (2) Which architecture turns generation data into transferable supervision beyond additional data, capacity, or training cost? (3) Which generation tasks contribute most to understanding gains? (4) At which layer should generation supervision be injected? (5) Which representation-level changes accompany the downstream gains?

\subsection{Generalization Across Training Stages}
\label{sec:generalization}

Firstly, we verify that our generation-guided framework is not limited to a specific training stage or backbone quality by applying GAS training atop understanding backbones of varying capability levels. We evaluate three settings: (i) training from scratch (the raw LLM backbone), (ii) after pretraining with large-scale multimodal data, and (iii) after directly supervised fine-tuning (SFT) based on Qwen3-VL-2B~\cite{baiQwen3VLTechnicalReport2025}. Each setting compares GAS with its corresponding standard VLM baseline; matched controls for generation-data exposure and additional training budget are reported separately in Section~\ref{sec:conflict}. Results are presented in Table~\ref{tab:generalization}.

\begin{table}[htbp]
    \renewcommand{\arraystretch}{0.98}
    \centering
    \setlength{\tabcolsep}{3pt}
    \resizebox{\textwidth}{!}{
    \begin{tabular}{@{}llccccc|cccc@{}}
        \toprule
        \textbf{Stage} & \textbf{Configuration} & \textbf{Overall} & \textbf{Perception} & \textbf{Reasoning} & \textbf{Count\&Spatial} & \textbf{Video} & \textbf{MME} & \textbf{MMMU} & \textbf{MathVista} & \textbf{MVBench} \\
        \midrule
        \multirow{2}{*}{From Scratch} & Baseline & 47.25 & 56.55 & 31.38 & 76.92 & 46.60 & 1848 & 43.5 & 42.7 & 46.9 \\
        & + GAS & \textbf{48.25} & \textbf{58.52} & \textbf{32.63} & 75.15 & \textbf{47.65} & \textbf{1895} & \textbf{45.8} & \textbf{47.1} & \textbf{49.5} \\
        \midrule
        \multirow{2}{*}{After Pretrain} & Baseline & 56.23 & 63.80 & 43.22 & 82.40 & 53.95 & 2169 & 53.2 & 72.3 & 55.1 \\
        & + GAS & \textbf{56.71} & \textbf{63.88} & \textbf{44.12} & \textbf{82.95} & 53.95 & 2170 & 51.1 & 71.2 & 54.6 \\
        \midrule
        \multirow{2}{*}{After SFT} & Baseline & 52.65 & 61.32 & 38.23 & 79.82 & 51.40 & 2060 & 46.3 & 52.9 & 53.0 \\
        & + GAS & \textbf{53.49} & \textbf{61.62} & \textbf{39.57} & \textbf{80.80} & \textbf{51.65} & \textbf{2096} & 45.8 & \textbf{53.6} & 52.9 \\
        \bottomrule
        
    \end{tabular}
    }
    \caption{\textbf{Generalization across training stages.} GAS generation-guided training yields improvements when applied to understanding backbones at different capability levels. }
    \label{tab:generalization}
\end{table}

The from-scratch setting yields the largest Overall gains ($+1.00$pp overall, with $+1.97$pp on Perception and $+1.25$pp on Reasoning), but a decrease on Count\&Spatial ($-1.77$pp), as the model benefits most from generation supervision when its visual representations are not yet well-developed. 
After large-scale pretraining, GAS still provides consistent improvements ($+0.48$pp overall, $+0.90$pp on Reasoning), showing that generation-guided supervision is complementary even to a strong, well-trained backbone. Most notably, applying GAS during SFT also yields a robust $+0.84$pp overall gain with simultaneous improvements on Perception ($+0.30$pp), Reasoning ($+1.34$pp), Count\&Spatial ($+0.98$pp), and Video ($+0.25$pp). 
Across three stages spanning the full spectrum of backbone capability, GAS delivers positive transfer on virtually every dimension, confirming that the proposed generation-guided training is a stage-agnostic augmentation that benefits both weak and already-strong understanding models.

\begin{tcolorbox}[colback=gray!8, colframe=gray!50, boxrule=0.5pt, arc=3mm, left=6pt, right=6pt, top=4pt, bottom=4pt]
\textbf{Insight 1:} Generation-guided training is stage-agnostic: it provides the largest gains from scratch, complementary improvements after pretraining, and still consistent benefits after SFT, making it a plug-in augmentation across the full training pipeline rather than a stage-specific technique.
\end{tcolorbox}

\subsection{What Makes Generation Supervision Transfer?}
\label{sec:conflict}

A generation-augmented model changes several factors at once: it sees a new data stream, spends additional training compute, introduces a generation branch, and learns from a new visual objective. We disentangle these factors with the controls in Table~\ref{tab:conflict}. Beyond the standard 2.5B-token understanding baseline, we train a compute-matched baseline with 11\% more understanding data and expose a standard VLM baseline to the same generation samples, whose image outputs contribute no loss. We then isolate the MoT architecture without generation supervision, apply NEP directly to a shared backbone, and replace NEP with position-aligned visual prediction (no shift) while keeping MoT and the training mixture fixed.

\begin{table}[htbp]
    \renewcommand{\arraystretch}{0.98}
    \centering
    \setlength{\tabcolsep}{3pt}
    \resizebox{\textwidth}{!}{
    \begin{tabular}{@{}lccccc@{}}
        \toprule
        \textbf{Configuration} & \textbf{Overall} & \textbf{Perception} & \textbf{Reasoning} & \textbf{Count\&Spatial} & \textbf{Video} \\
        \midrule
        Understanding-only baseline & 47.25 & 56.55 & 31.38 & \textbf{76.92} & 46.60 \\
        Baseline, +11\% understanding data & 47.73 & 57.63 & 34.01 & 71.28 & 45.55 \\
        Baseline, same generation data (text loss only) & 47.14 & 56.08 & 33.52 & 71.00 & 46.25 \\
        \midrule
        MoT, no generation supervision & 47.63 & 57.22 & \textbf{34.20} & 70.78 & 45.60 \\
        Shared backbone + NEP & 46.00 & 54.95 & 32.25 & 70.20 & 45.15 \\
        MoT + no-shift visual loss & 47.84 & 57.70 & 32.16 & 76.38 & 46.60 \\
        MoT + NEP (GAS) & \textbf{48.25} & \textbf{58.52} & 32.63 & 75.15 & \textbf{47.65} \\
        \bottomrule
    \end{tabular}
    }
    \caption{\textbf{Disentangling data, architecture, and objective.} Matched controls separate additional training, passive exposure to generation data, MoT decoupling, and the NEP objective.}
    \label{tab:conflict}
\end{table}

Additional training alone does not explain the gain. The longer understanding-only run improves Overall only modestly, while merely mixing in the same generation samples provides no benefit because their image targets are not supervised by standard VLM training. MoT without a generation loss also improves some dimensions but does not match GAS, showing that neither data exposure nor the extra branch is sufficient by itself.

The objective and its gradient route are complementary. Applying NEP directly to the shared upper layers causes broad degradation, revealing interference between generation-specific optimization and the understanding pathway. With MoT fixed, position-aligned no-shift prediction already provides useful visual supervision, whereas autoregressive NEP yields the strongest aggregate transfer and the best Perception and Video scores. The no-shift control remains stronger on Count\&Spatial, so the effect is not a uniform per-capability improvement; rather, NEP changes the transfer profile while improving the aggregate outcome. Together, these controls show that MoT protects the deployed understanding pathway and NEP supplies a task-conditioned learning signal that can enrich their shared lower representation.

\begin{tcolorbox}[colback=gray!8, colframe=gray!50, boxrule=0.5pt, arc=3mm, left=6pt, right=6pt, top=4pt, bottom=4pt]
\textbf{Insight 2:} Generation data is not supervision by itself. GAS turns it into transferable visual learning by pairing task-conditioned next-embedding prediction with an isolated upper generation pathway: NEP enriches the shared representation, while MoT prevents generation-specific optimization from rewriting the understanding model.
\end{tcolorbox}

\subsection{Per-Task Generation Contribution}
\label{sec:pertask}

We next investigate which generation tasks contribute most effectively to understanding enhancement by individually adding 200k samples from each of the five task categories to the understanding-only baseline. Table~\ref{tab:pertask} presents the results, and we additionally report the combined ``All'' task samples configuration.

Several patterns emerge from this ablation. First, all five generation tasks improve at least one evaluation dimension, but their contribution profiles differ significantly, validating our latent task correlation hypothesis (Section~\ref{sec:task_construction}). 
T2I provides the highest single-task overall gain (+0.49pp) with broad improvements across perception and spatial understanding; Segmentation and Grounding strongly boost Count\&Spatial (+2.00 and +2.37pp respectively, with CV-Bench 3D improving by up to 9.8pp from segmentation), consistent with the expectation that pixel-level and region-level generation tasks directly cultivate spatial perception capabilities. 
Visual-CoT uniquely enhances complex reasoning benchmarks: BLINK (+4.2pp) and MathVision (+1.0pp), reflecting its role in forcing the model to produce intermediate visual reasoning traces that reinforce multi-step inference. Editing yields a small Overall gain and a larger Count\&Spatial improvement, but does not improve every dimension. 

\begin{table}[htbp]
    \renewcommand{\arraystretch}{0.98}
    \centering
    \setlength{\tabcolsep}{3pt}
    \resizebox{\textwidth}{!}{
    \begin{tabular}{@{}lccccc|ccccc@{}}
        \toprule
        \textbf{Configuration} & \textbf{Overall} & \textbf{Perception} & \textbf{Reasoning} & \textbf{Count\&Spatial} & \textbf{Video} & \textbf{MME} & \textbf{BLINK} & \textbf{CV-2D/3D} & \textbf{MathVision} & \textbf{VisuLogic} \\
        \midrule
        Understanding only & 47.63 & 57.22 & 34.20 & 70.78 & 45.60 & 1868 & 41.9 & 59.8/52.1 & 17.1 & 25.2 \\
        \midrule
        + T2I & 48.12 & 57.70 & 34.66 & 71.65 & 45.85 & 1820 & 44.5 & 59.3/55.3 & 15.8 & 25.1 \\
        + Editing & 47.77 & 56.70 & 34.12 & 72.95 & 45.70 & 1830 & 42.5 & 61.1/59.5 & 14.8 & 26.4 \\
        + Segmentation & 47.43 & 55.32 & 34.35 & 72.78 & 45.55 & 1750 & 44.1 & 63.6/55.9 & 16.4 & 24.2 \\
        + Grounding & 47.92 & 57.28 & 33.92 & 73.15 & 45.95 & 1840 & 44.6 & 61.9/58.3 & 13.5 & 23.6 \\
        + Visual-CoT & 47.86 & 57.80 & 34.26 & 71.45 & 45.20 & 1877 & 46.1 & 56.9/57.3 & 18.1 & 26.6 \\
        \midrule
        + All (combined) & \emph{48.30} & 57.08 & 34.28 & \textbf{75.72} & 45.40 & 1851 & \emph{44.9} & \textbf{62.5/67.0} & \textbf{20.1} & 23.6 \\
        \bottomrule
    \end{tabular}
    }
    \caption{\textbf{Per-task ablation.} Each row adds a single generation task category to the understanding-only baseline. The ``All'' row combines all five categories. }
    \label{tab:pertask}
\end{table}

Most notably, the ``All'' configuration achieves a Count\&Spatial score of 75.72, higher than any single-task setting and exceeding the strongest individual contributors (Segmentation $+2.00$pp, Grounding $+2.37$pp), with a similar pattern observed on MathVision (20.1 in the combined setting versus 18.1 from the best single task). This pattern is suggestive of a \emph{complementary} effect: diverse generation tasks appear to address different facets of fine-grained visual representation that no single category fully covers. 
The practical implication is the one we use throughout the paper: composing tasks with distinct latent correlations to understanding is a productive design lever, especially for perception- and counting-oriented capabilities.

\paragraph{Effect of Task Correlation on Understanding Gains.}

\begin{wraptable}{r}{0.46\linewidth}
    \vspace{-0.7\baselineskip}
    \centering
    \captionsetup{
        font=small,
        skip=3pt,
        width=\linewidth,
        justification=raggedright,
        singlelinecheck=false
    }
    \footnotesize
    \renewcommand{\arraystretch}{1.02}

    \begin{tabular*}{\linewidth}{@{\extracolsep{\fill}}lccc@{}}
        \toprule
        \textbf{Metric}
        & \textbf{Baseline}
        & \textbf{Easy}
        & \textbf{Rewrite} \\
        \midrule
        Overall
        & 47.63 & 47.87 & \textbf{48.34} \\
        Perception
        & 57.22 & 56.47 & \textbf{57.53} \\
        Reasoning
        & 34.20 & 34.37 & \textbf{34.58} \\
        Count\&Spatial
        & 70.78 & 72.47 & \textbf{73.62} \\
        Video
        & 45.60 & \textbf{46.55} & 45.95 \\
        \bottomrule
    \end{tabular*}

    \caption[T2I prompt correlation.]{
        \textbf{Effect of T2I prompt correlation with understanding tasks.} 
        Aligning T2I prompts with the compositional, knowledge-grounded scene structures required by understanding benchmarks improves transfer at fixed task and data volume.
    }
    \label{tab:t2i_complexity}
    \vspace{-0.6\baselineskip}
\end{wraptable}

The heterogeneous gains across task categories suggest that transfer depends not only on task modality, but also on the latent correlation between generation signal and the target understanding capability, as well as on the internal characteristics of the generation task itself. 
To probe this factor more directly, we conduct a controlled study by fixing the T2I task and data volume and varying only prompt characteristics.
Specifically, we collect 200k additional T2I samples from COYO-700M~\cite{kakaobrain2022coyo-700m} dataset and compare two variants: (i) \textbf{Easy}, generic short-caption synthesis as commonly used in prior unified models, and (ii) \textbf{Rewrite}, where prompts are enriched with world knowledge, multi-object compositions, and explicit attribute/relation cues to better mirror the scene structures probed by our reasoning and perception benchmarks.

As Table~\ref{tab:t2i_complexity} shows, generic T2I yields a small Overall gain ($+0.24$pp) but decreases Perception by $0.75$pp. Rewritten prompts improve Overall by $+0.71$pp and all four capability aggregates relative to the same baseline, although the easy prompts retain the best Video score. 
This contrast shows that volume of generation supervision is not what matters, a poorly correlated generation task can act as a distractor rather than a regularizer. 
Once prompts are rewritten to demand the same compositional, knowledge-grounded scene structures probed by understanding benchmarks, the very same T2I category turns into a broadly beneficial signal, on par with our targeted tasks. This directly closes the loop with our latent task correlation principle (Section~\ref{sec:task_construction}): even within one nominal task, transfer scales with task-level correlation, not with sample count.

\begin{tcolorbox}[colback=gray!8, colframe=gray!50, boxrule=0.5pt, arc=3mm, left=6pt, right=6pt, top=4pt, bottom=4pt]
\textbf{Insight 3:} Understanding gains scale with task correlation. Each task targets a distinct facet, like spatial tasks for counting/localization, Visual-CoT for reasoning, editing for holistic perception, and combining them tends to yield further, complementary improvements on perception- and counting-oriented benchmarks beyond any single category. Even within T2I, correlation-aware prompts unlock transfer that generic captions cannot.

\end{tcolorbox}

\subsection{Layer-wise Supervision Injection}
\label{sec:layer}

Our MoT architecture extracts features from an intermediate layer $l_s$ of the understanding backbone to feed into the generation branch. This design choice raises the question: at which depth should generation supervision be injected? We compare extracting hidden states from layers 8, 14 (approximately $L/2$ for our 2B, 28-layers model), and 20, as well as the effect of progressive loss warmup and unfreezing the ViT encoder. Results are shown in Table~\ref{tab:layer}.

\begin{table}[htbp]
    \renewcommand{\arraystretch}{0.98}
    \centering
    \setlength{\tabcolsep}{3pt}
    \resizebox{\textwidth}{!}{
    \begin{tabular}{@{}lccccc|cccc@{}}
        \toprule
        \textbf{Configuration} & \textbf{Overall} & \textbf{Perception} & \textbf{Reasoning} & \textbf{Count\&Spatial} & \textbf{Video} & \textbf{MME} & \textbf{BLINK} & \textbf{CountBenchQA} & \textbf{CV-3D} \\
        \midrule
        Layer 8 & 47.06 & 56.80 & 31.52 & 75.88 & 45.40 & 1830 & 44.3 & 86.4 & 66.1 \\
        Layer 14 & 47.85 & 58.58 & 31.64 & 75.28 & 47.60 & 1849 & 47.7 & 86.4 & 64.6 \\
        \textbf{Layer 14 + warmup} & \textbf{48.25} & \textbf{58.52} & \textbf{32.63} & 75.15 & \textbf{47.65} & \textbf{1895} & 45.8 & \textbf{87.7} & 63.6 \\
        Layer 14 + unfreeze ViT & 46.20 & 55.80 & 31.31 & 72.60 & 45.25 & 1808 & 44.7 & 79.7 & 69.9 \\
        Layer 20 & 47.25 & 58.20 & 31.43 & 74.60 & 45.45 & 1855 & 44.0 & 86.0 & 64.2 \\
        \bottomrule
    \end{tabular}
    }
    \caption{\textbf{Layer-wise injection and training strategy ablation.} We vary the layer from which hidden states are extracted for the generation branch, and compare with progressive warmup and ViT unfreezing. Layer 14 ($\approx L/2$) with progressive warmup achieves the best overall performance.}
    \label{tab:layer}
\end{table}

The intermediate layer 14 (approximately $L/2$) consistently outperforms both the shallower layer~8 and the deeper layer~20 on overall metrics and perception, while layer~8 achieves the strongest Count\&Spatial score (75.88), suggesting that early-layer features already encode rich spatial information that benefits from generation supervision. Layer~14 achieves the highest BLINK (47.7/45.8), indicating that deeper features carry more abstract visual-semantic information beneficial for multi-image reasoning, represents a balance point where features retain sufficient spatial detail while having undergone enough semantic abstraction to benefit broad understanding tasks.

Progressive loss warmup improves the raw layer-14 configuration by $+0.40$pp Overall and $+0.99$pp Reasoning, consistent with our motivation that the faster-converging generation loss should be introduced gradually. 
Conversely, unfreezing the ViT encoder reduces Overall by $1.65$pp and CountBenchQA from 86.4 to 79.7 relative to the frozen layer-14 setting, indicating that the generation-side optimization pressure corrupts the pretrained visual features when allowed to propagate into the encoder, which further validating our design to keep the ViT frozen and channel generation benefits exclusively through the shared projection layer.

\begin{tcolorbox}[colback=gray!8, colframe=gray!50, boxrule=0.5pt, arc=3mm, left=6pt, right=6pt, top=4pt, bottom=4pt]
\textbf{Insight 4:} The best Overall score occurs when supervision branches near $L/2$ with progressive warmup, balancing spatial detail and semantic abstraction. Unfreezing the ViT degrades performance in this setup, suggesting that generation benefits should flow through the projection layer, not corrupt the frozen encoder.
\end{tcolorbox}

\subsection{Representation-Level Diagnostics}
\label{sec:representation}

To understand how generation training improves understanding at the representation level, we conduct three complementary analyses: (i) visual information retention across layers, (ii) attention pattern visualization, and (iii) linear probing diagnostics.

\begin{figure*}[t]
    \centering
    \begin{minipage}[c]{0.56\textwidth}
        \centering
        \includegraphics[width=\linewidth]{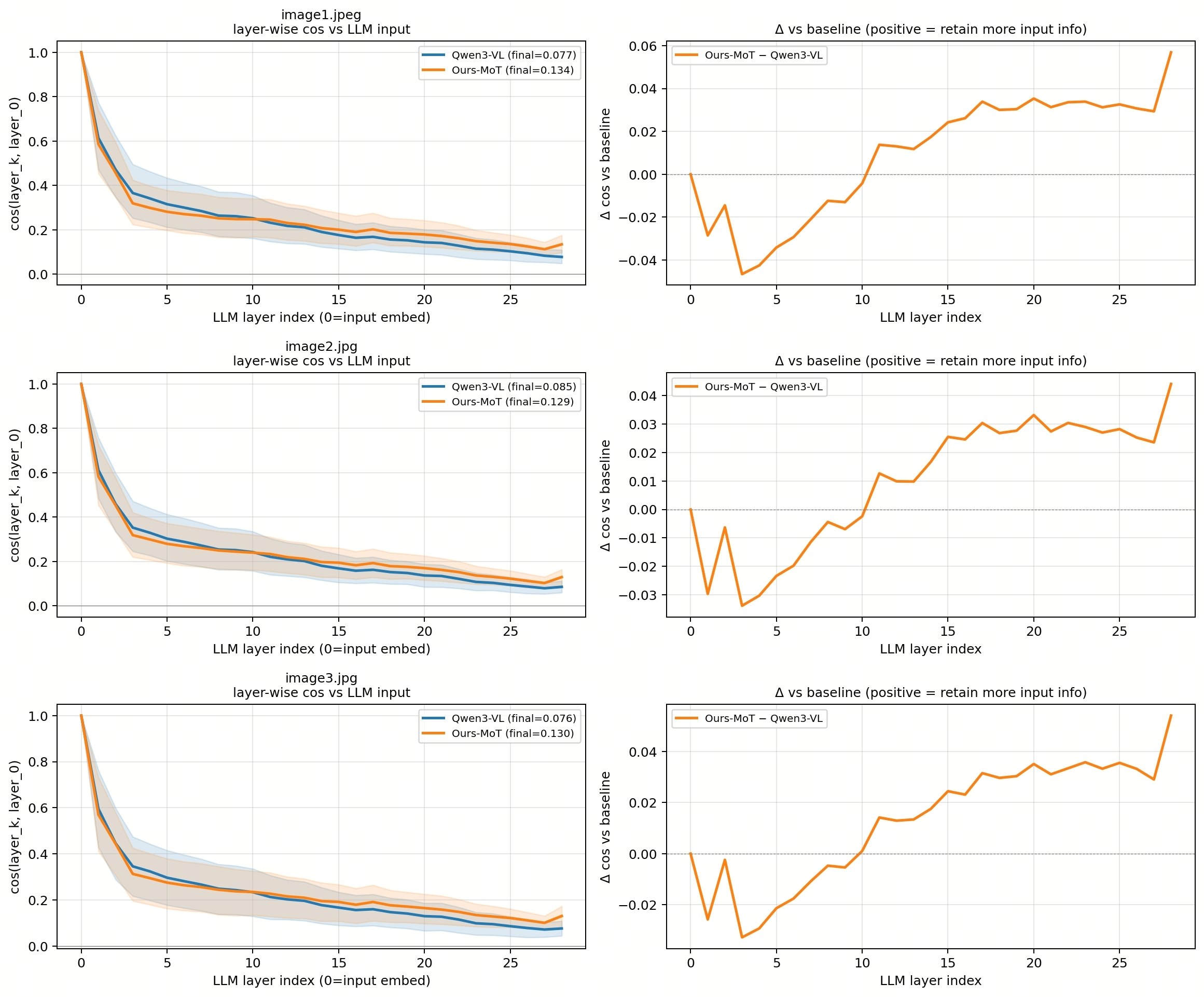}
    \end{minipage}\hfill
    \begin{minipage}[c]{0.42\textwidth}
        \centering
        \includegraphics[width=\linewidth]{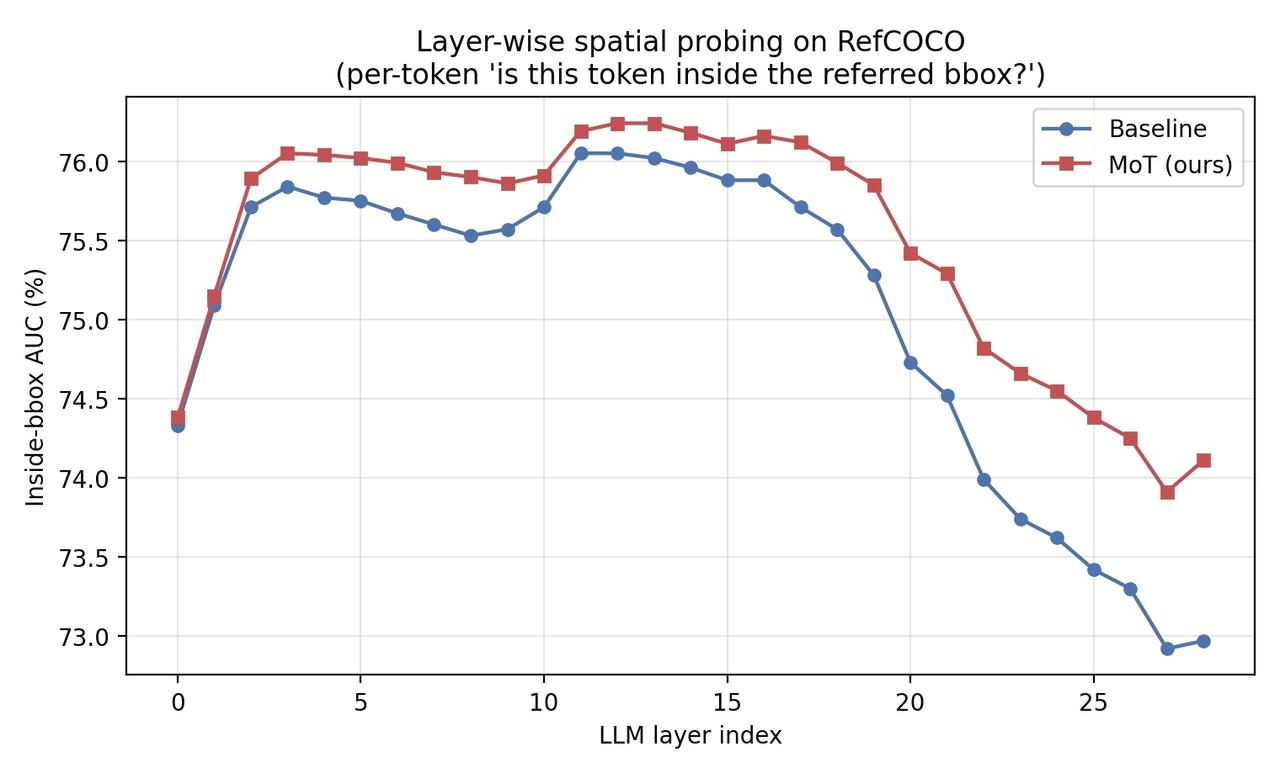}\\[2pt]
        \includegraphics[width=\linewidth]{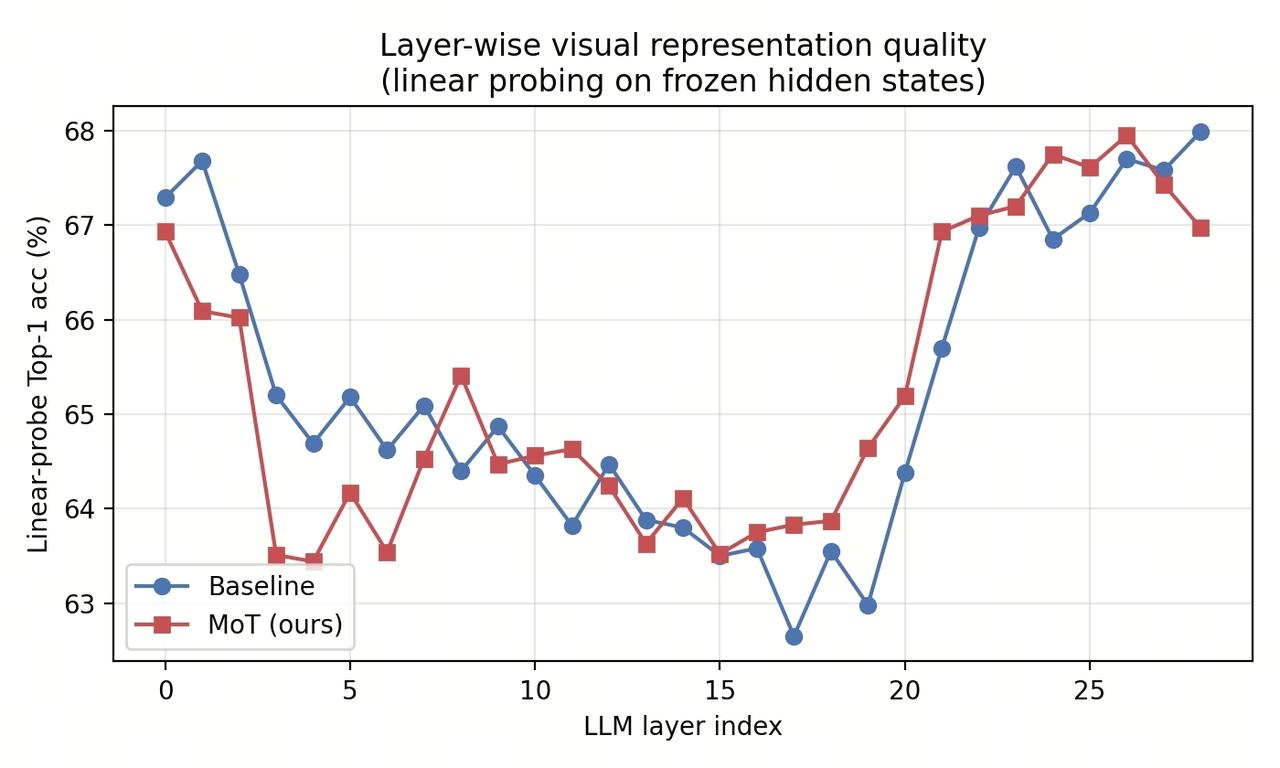}
    \end{minipage}
    \caption{\textbf{Visual information retention and layer-wise linear probing.} \emph{Left:} cosine similarity between per-layer visual tokens and input visual features.
    \emph{Right top / bottom:} linear probing accuracy on RefCOCO (region-level) and ImageNet (global). 
    The models are comparable on ImageNet, while GAS's representations become increasingly more linearly separable on RefCOCO in deeper layers.
    }
    \label{fig:retention}
\end{figure*}

\paragraph{Visual Information Retention.}
We measure the cosine similarity between the visual token representations at each LLM layer and the original visual features at the LLM input, tracking how much visual information is preserved as processing deepens. 
Figure~\ref{fig:retention} (left) shows that the baseline similarity decreases beyond layer $L/3$ whereas the GAS model retains higher similarity through the middle and deep layers, demonstrating that generation supervision encourages the network to retain visual features rather than allowing them to be ``squeezed out'' by language-dominant processing. The generation branch, by requiring accurate embedding prediction from intermediate representations, creates an implicit incentive for the backbone to preserve visual information at those layers, which then propagates benefit to the understanding branch operating on the same representations.

\paragraph{Layer-wise Linear Probing.}
To quantify the discriminative quality of visual representations at each layer, we extract hidden states corresponding to visual tokens and train linear classifiers on ImageNet~\cite{deng2009imagenet} (global classification) and RefCOCO~\cite{mao2016generation,kazemzadeh2014referitgame} (region-level referring expression comprehension). As shown in Figure~\ref{fig:retention} (right), the two models perform comparably on ImageNet across all layers. 
On RefCOCO, GAS exhibits a growing advantage that widens consistently from layer 8 onward, ultimately achieving substantially higher probing accuracy at the final layer. This pattern reveals that generation training specifically enriches \emph{fine-grained, region-level} visual representations rather than generic semantic features, aligning with GAS's strongest downstream gains on spatial and grounding evaluations. 
The widening gap at deeper layers further corroborates the retention analysis: because GAS preserves more spatial visual information in deep layers, its representations remain useful for fine-grained tasks even after extensive language-driven processing.

\begin{figure}[t]
    \centering
    \setlength{\tabcolsep}{2pt}
    \renewcommand{\arraystretch}{0.9}
    \begin{tabular}{c c c}
         & \textbf{Baseline} & \textbf{GAS} \\
        \rotatebox{90}{\textbf{CV-Bench}} &
        \includegraphics[width=0.42\textwidth]{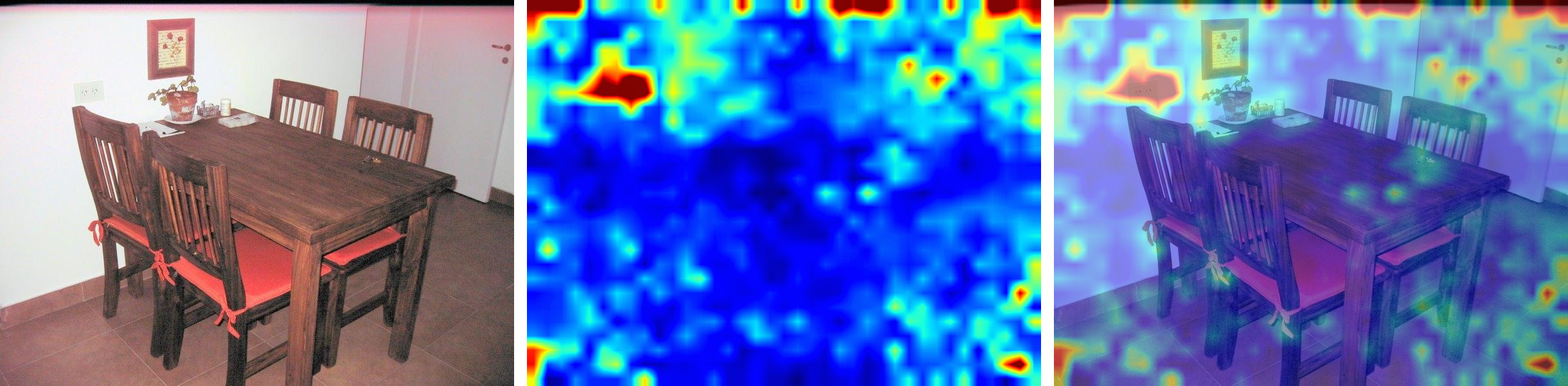} &
        \includegraphics[width=0.42\textwidth]{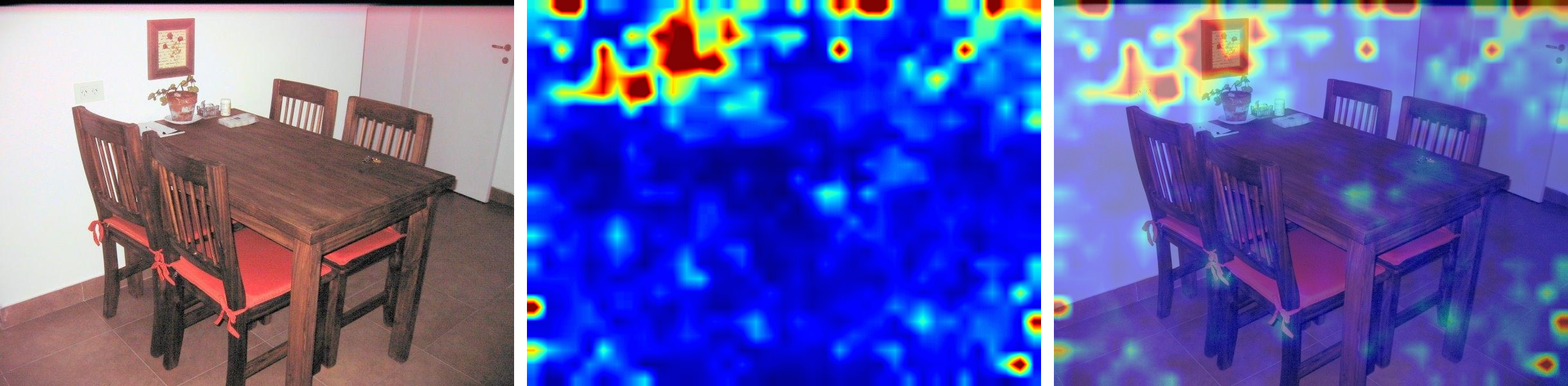} \\
        \rotatebox{90}{\textbf{MathVista}} &
        \includegraphics[width=0.42\textwidth]{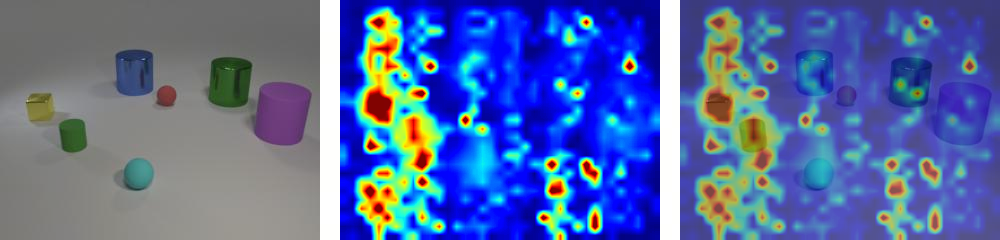} &
        \includegraphics[width=0.42\textwidth]{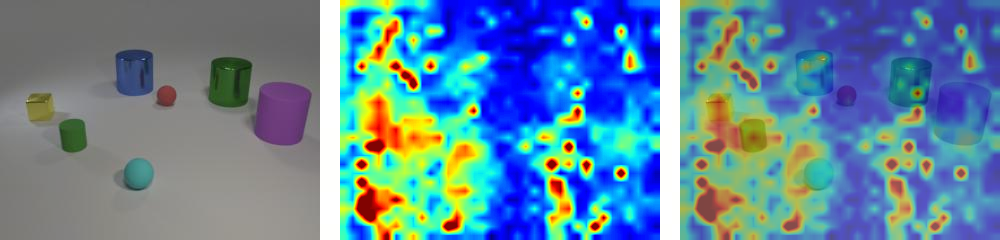} \\
        \rotatebox{90}{\textbf{~~~~~~Layer-wise Attention Map}} &
        \includegraphics[width=0.42\textwidth]{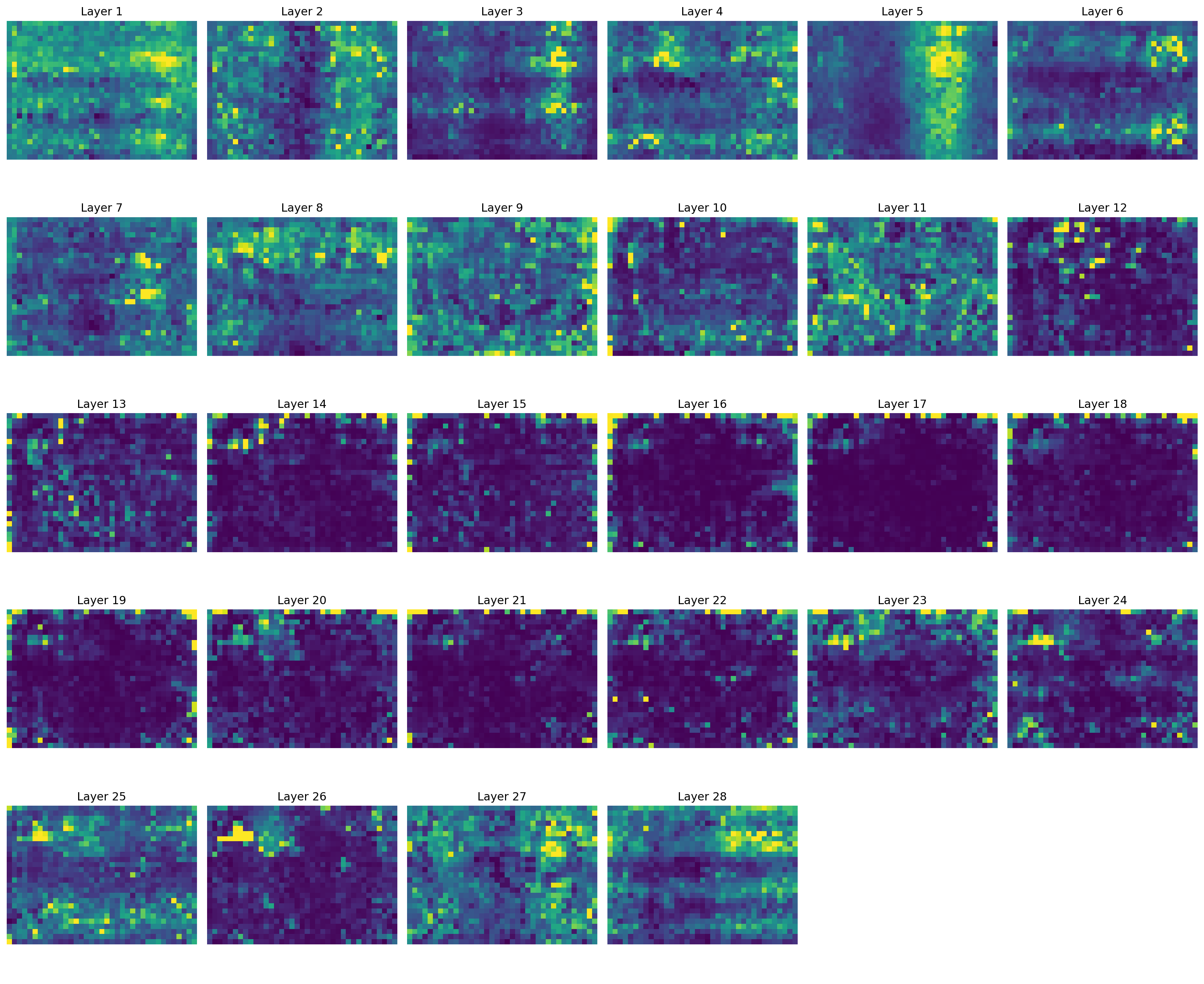} &
        \includegraphics[width=0.42\textwidth]{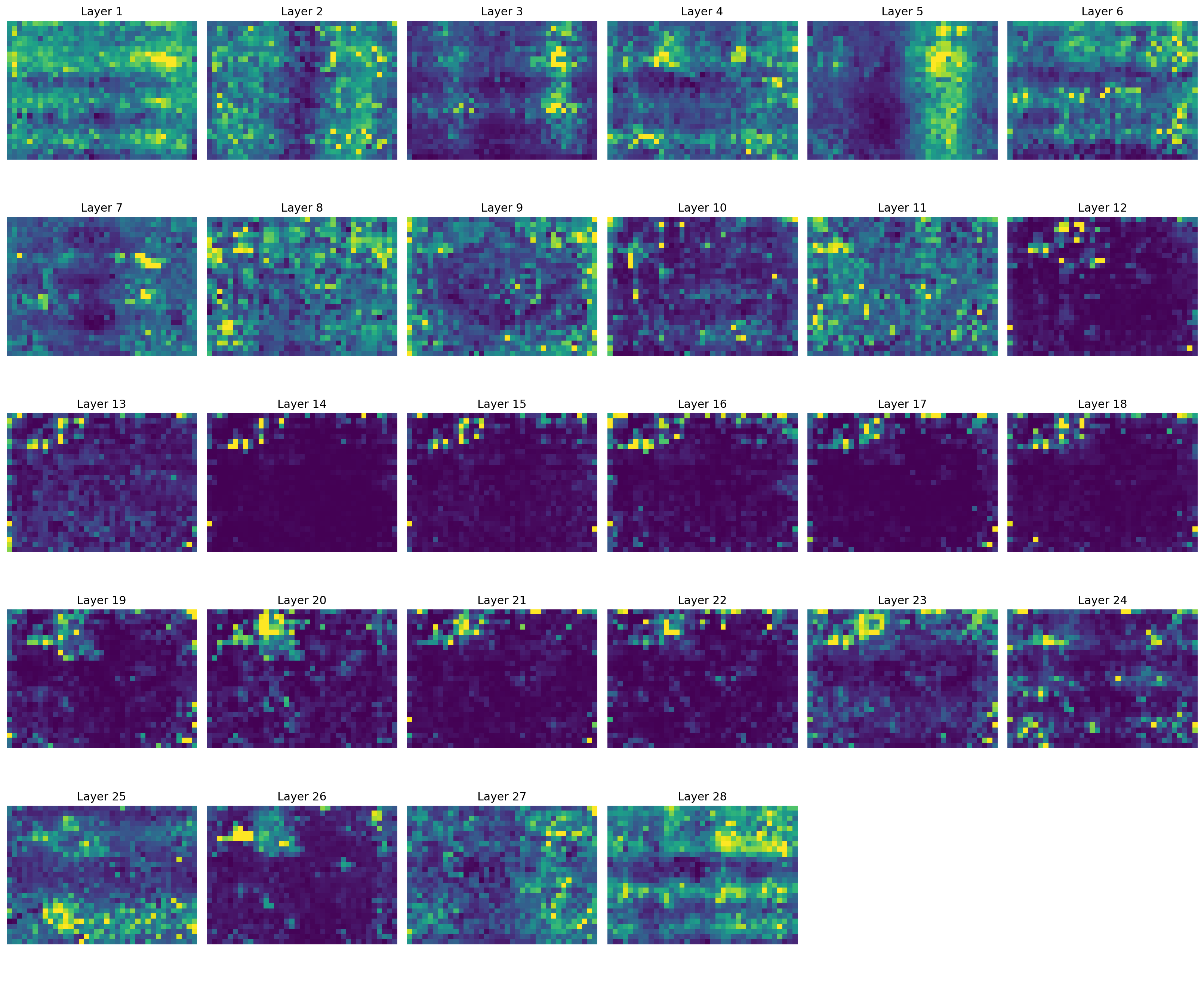} \\
    \end{tabular}
    \caption{\textbf{Layer-wise attention visualization on representative understanding samples.} The first two rows compare visual-token attention for the baseline and GAS model on CV-Bench-2D~\cite{tong2024cambrian} and MathVista~\cite{lu2024mathvista}. 
    The bottom row shows the full layer-by-layer attention evolution (layers 1--28): the baseline's attention rapidly diffuses and collapses across mid-to-deep layers, whereas the GAS model preserves sharp, task-relevant focus throughout the network depth and produces the corrected predictions shown in Boxes~\ref{box:cvbench_attn} and~\ref{box:mathvista_attn}.
    }
    \label{fig:attention}
    \vspace{-0.5em}
\end{figure}

\paragraph{Attention Visualization.}
We visualize attention maps from representative understanding CV-Bench and MathVista examples at different LLM layers to examine where the model ``looks'' during reasoning. 
As illustrated in Figure~\ref{fig:attention}, the baseline model's attention becomes more diffuse in deeper layers---by layer 20, attention is scattered broadly across the image with no clear focus on task-relevant regions. 
In contrast, GAS maintains sharp, focused attention on question-relevant regions throughout the network depth. 
For instance, on a CV-Bench counting question (``How many pictures are in the image?''), the baseline incorrectly answers 0 with diffuse attention, while our model correctly identifies the single picture with attention concentrated on the relevant region. 
On MathVista object-counting tasks, GAS correctly attends to all relevant objects for subtraction reasoning, whereas the baseline's scattered attention leads to miscounting. 
This attention sharpening directly explains the mechanism by which generation training improves understanding: by requiring the generation branch to predict precise visual embeddings, the shared representations develop stronger spatial selectivity that carries over to understanding-time inference.

\vspace{0.5em}
\refstepcounter{qualbox}\label{box:cvbench_attn}
\begin{tcolorbox}[qualbox, title = {Qualitative Case~\thequalbox: CV-Bench-2D (Attention Analysis)}]
\textbf{Question:} How many pictures are in the image? \quad A.~1 \quad B.~3 \quad C.~2 \quad D.~0 \\
\textbf{Ground Truth:} \textcolor{tablegreen}{A} \\[2pt]
\textbf{\textcolor{red}{Baseline:}}
\textbf{\textcolor{blue}{<think>}} The image shows a room with a wooden table and chairs; no separate pictures appear, so the answer is D (0). \textbf{\textcolor{blue}{</think>}}
Answer: \textbf{\textcolor{red}{D \ding{55}}} \\
\textbf{GAS:}
\textbf{\textcolor{blue}{<think>}} The image is itself a single coherent scene (a dining setup); among the choices, A (1) corresponds to the single picture present. \textbf{\textcolor{blue}{</think>}}
Answer: \textbf{\textcolor{tablegreen}{A \ding{51}}}
\end{tcolorbox}

\refstepcounter{qualbox}\label{box:mathvista_attn}
\begin{tcolorbox}[qualbox, title = {Qualitative Case~\thequalbox: MathVista-MINI (Attention Analysis)}]
\textbf{Question:} Subtract all red things. Subtract all tiny matte balls. How many objects are left? \\
\textbf{Ground Truth:} \textcolor{tablegreen}{5} \\[2pt]
\textbf{\textcolor{red}{Baseline:}}
\textbf{\textcolor{blue}{<think>}} Identifies the red matte ball as both ``red'' and ``tiny matte ball'', but then double-counts the remaining items, mistakenly listing 6 leftover objects. \textbf{\textcolor{blue}{</think>}}
Answer: \textbf{\textcolor{red}{6 \ding{55}}} \\
\textbf{GAS:}
\textbf{\textcolor{blue}{<think>}} Total 7 objects; the red sphere is the only object that is both red and a tiny matte ball, so it is removed once. Remaining objects: yellow cube, green small cylinder, cyan sphere, blue / green / purple large cylinders, giving $7-1-1=5$. \textbf{\textcolor{blue}{</think>}}
Answer: \textbf{\textcolor{tablegreen}{5 \ding{51}}}
\end{tcolorbox}

\begin{tcolorbox}[colback=gray!8, colframe=gray!50, boxrule=0.5pt, arc=3mm, left=6pt, right=6pt, top=4pt, bottom=4pt]
\textbf{Insight 5:} Generation training enhances understanding through three complementary representational mechanisms: (i) improved visual information retention in deep layers, preventing the progressive attenuation observed in text-only training; (ii) enriched fine-grained spatial representations (validated by linear probing on RefCOCO); and (iii) sharper, task-focused attention patterns that persist throughout the reasoning chain.
\end{tcolorbox}

\paragraph{Text-only Reasoning capability Mildly Improved.}
A natural concern is whether generation supervision harms the LLM backbone's pure-text reasoning ability. We compare GAS with the understanding-only baseline on \textbf{ZebraLogic}~\cite{lin2025zebralogic} (constraint-satisfaction logic puzzles) and \textbf{MMLU-Redux}~\cite{gema2025we} (a re-annotated, less noisy subset of MMLU). 
Table~\ref{tab:textonly} shows higher point estimates for GAS on both benchmarks: ZebraLogic puzzle accuracy increases by $+2.8$pp (with consistent improvements on small/medium puzzles and cell-level accuracy), and MMLU-Redux improves by $+1.23$pp. 
This indicates that generation-guided multimodal training does not act as a regularizer that erodes language reasoning; on the contrary, the joint optimization mildly benefits language-only capability, likely because the visual generation branch absorbs visual-specific gradient pressure that would otherwise distort the LLM's text representations.

\begin{table}[htbp]
    \centering
    \footnotesize
    \setlength{\tabcolsep}{4pt}
    \renewcommand{\arraystretch}{0.95}
    \begin{tabular}{@{}lcccc@{}}
        \toprule
        \multirow{2}{*}{\textbf{Model}} & \multicolumn{3}{c}{\textbf{ZebraLogic}} & \textbf{MMLU-Redux} \\
        \cmidrule(lr){2-4} \cmidrule(lr){5-5}
         & \textbf{Puzzle Acc} & \textbf{Cell Acc} & \textbf{Small Acc} & \textbf{Acc} \\
        \midrule
        Qwen3-VL baseline                & 18.0 & 12.74 & 52.19 & 52.66 \\
        + GAS (Ours)                     & \textbf{20.8} & \textbf{14.79} & \textbf{56.56} & \textbf{53.89} \\
        \bottomrule
    \end{tabular}
    \caption{\textbf{Text-only reasoning evaluation.} GAS preserves and mildly improves pure-text reasoning on ZebraLogic and MMLU-Redux, indicating that generation supervision does not interfere with the LLM's language capability.}
    \label{tab:textonly}
\end{table}

\section{Related Works}
\label{sec:related_works}

Multimodal large language models (MLLMs) have emerged as the prevailing paradigm for image and video understanding, typically by aligning pretrained visual encoders with powerful language backbones.
Representative early systems include Flamingo \cite{alayrac2022flamingo} and InstructBLIP \cite{dai2023instructblip}, whereas subsequent open-source families such as the LLaVA series \cite{li2024llava,liuVisualInstructionTuning2023,li2024llavanext} and Qwen-VL series \cite{baiQwen3VLTechnicalReport2025,baiQwen25VLTechnicalReport2025} have further advanced image perception and understanding.
Most studies in this line largely follow the LLaVA paradigm \cite{liuVisualInstructionTuning2023}, in which visual inputs are first encoded by a vision encoder \cite{radford2021learning,tschannen2025siglip} and subsequently concatenated with text tokens for joint modeling by a language model decoder.
Proprietary systems such as GPT \cite{openaiGPT5} and Gemini \cite{IntroducingGemini202024,gemini3flash} likewise exhibit strong multimodal understanding capabilities.
Recent advances have further extended these models to multimodal visual reasoning \cite{xu2026visulogic,wang2026vgr,ke2025explain} and video understanding \cite{liu2025videoxlproreconstructivetokencompression,shu2025video,li2026videochatflash}.
Despite their notable strengths in semantic abstraction and cross-modal alignment, these models are primarily optimized for text generation rather than learning from native visual information. The training signal flows exclusively from vision to language -- visual tokens serve as a conditioning context for text prediction, but no supervisory gradient encourages the model to develop fine-grained, pixel-level visual perception or to retain visual evidence throughout deep reasoning chains. This asymmetry limits performance on tasks that demand precise spatial understanding, such as grounding, segmentation, and complex visual chain-of-thought reasoning.

\begin{table}[htbp]
    \centering
    \caption{\textbf{Design comparison of representative approaches that combine visual understanding and generation.} The attributes characterize how generation is used during training rather than absolute model quality. GAS is the only compared design that combines all six properties.}
    \label{tab:method_design_comparison}
    \scriptsize
    \setlength{\tabcolsep}{2.6pt}
    \renewcommand{\arraystretch}{1.13}
    \begin{tabularx}{0.995\textwidth}{@{}l X C{1.20cm} C{1.20cm} C{1.20cm} C{1.20cm} C{1.20cm} C{1.20cm}@{}}
        \toprule
        \textbf{Method} & \textbf{Generation target} & \makecell{\textbf{U-}\\\textbf{centric}} & \makecell{\textbf{Task-}\\\textbf{cond.}} & \makecell{\textbf{Exact}\\\textbf{U-space}} & \makecell{\textbf{No ext.}\\\textbf{codec}} & \makecell{\textbf{Upper}\\\textbf{U-isol.}} & \makecell{\textbf{Train-}\\\textbf{only aux.}} \\
        \midrule
        \multicolumn{8}{@{}l}{\emph{Native unified understanding--generation models}} \\
        Chameleon~\cite{team2024chameleon} & Discrete image tokens & \textcolor{red}{\xmark} & \textcolor{red}{\xmark} & \textcolor{red}{\xmark} & \textcolor{red}{\xmark} & \textcolor{red}{\xmark} & \textcolor{red}{\xmark} \\
        Show-o~\cite{xie2025showo} & Discrete diffusion tokens & \textcolor{red}{\xmark} & \textcolor{tablegreen}{\cmark} & \textcolor{red}{\xmark} & \textcolor{red}{\xmark} & \textcolor{red}{\xmark} & \textcolor{red}{\xmark} \\
        Janus-Pro~\cite{chen2025janus} & Discrete image tokens & \textcolor{red}{\xmark} & \textcolor{red}{\xmark} & \textcolor{red}{\xmark} & \textcolor{red}{\xmark} & \textcolor{red}{\xmark} & \textcolor{red}{\xmark} \\
        MetaMorph~\cite{tong2025metamorph} & Vision-encoder features & \textcolor{orange}{\pmark} & \textcolor{tablegreen}{\cmark} & \textcolor{red}{\xmark} & \textcolor{red}{\xmark} & \textcolor{red}{\xmark} & \textcolor{red}{\xmark} \\
        BAGEL~\cite{deng2025emerging} & VAE/semantic features & \textcolor{red}{\xmark} & \textcolor{tablegreen}{\cmark} & \textcolor{red}{\xmark} & \textcolor{red}{\xmark} & \textcolor{orange}{\pmark} & \textcolor{red}{\xmark} \\
        UniFork~\cite{li2025unifork} & Discrete image tokens & \textcolor{red}{\xmark} & \textcolor{red}{\xmark} & \textcolor{red}{\xmark} & \textcolor{red}{\xmark} & \textcolor{tablegreen}{\cmark} & \textcolor{red}{\xmark} \\
        \midrule
        \multicolumn{8}{@{}l}{\emph{Generation supervision designed to enhance understanding}} \\
        ROSS~\cite{wang2025reconstructive} & Continuous VAE features & \textcolor{tablegreen}{\cmark} & \textcolor{red}{\xmark} & \textcolor{red}{\xmark} & \textcolor{red}{\xmark} & \textcolor{red}{\xmark} & \textcolor{tablegreen}{\cmark} \\
        ASVR~\cite{wang2026autoregressive} & Discrete semantic tokens & \textcolor{tablegreen}{\cmark} & \textcolor{red}{\xmark} & \textcolor{red}{\xmark} & \textcolor{red}{\xmark} & \textcolor{red}{\xmark} & \textcolor{tablegreen}{\cmark} \\
        UniHetero~\cite{chen2025unihetero} & LLM-input embeddings & \textcolor{tablegreen}{\cmark} & \textcolor{red}{\xmark} & \textcolor{tablegreen}{\cmark} & \textcolor{red}{\xmark} & \textcolor{red}{\xmark} & \textcolor{orange}{\pmark} \\
        UniMRG~\cite{su2026generation} & RGB/depth/segmentation & \textcolor{tablegreen}{\cmark} & \textcolor{tablegreen}{\cmark} & \textcolor{red}{\xmark} & \textcolor{red}{\xmark} & \textcolor{orange}{\pmark} & \textcolor{red}{\xmark} \\
        \midrule
        \rowcolor{lightgray}
        \textbf{GAS (Ours)} & \textbf{LLM-input embeddings} & \textcolor{tablegreen}{\cmark} & \textcolor{tablegreen}{\cmark} & \textcolor{tablegreen}{\cmark} & \textcolor{tablegreen}{\cmark} & \textcolor{tablegreen}{\cmark} & \textcolor{tablegreen}{\cmark} \\
        \bottomrule
        \vspace{0.5pt}
    \end{tabularx}
    \vspace{2pt}
    \begin{minipage}{0.99\textwidth}
    \scriptsize
    \textit{Definitions.} \textbf{U-centric}: generation is explicitly optimized to improve understanding rather than retained as a co-equal capability; \textbf{Task-cond.}: an observed visual input and instruction determine a distinct target, beyond generic text-to-image synthesis or input reconstruction; \textbf{Exact U-space}: the target is the projected continuous embedding consumed by the understanding LLM; \textbf{No ext. codec}: no separate visual tokenizer, VAE, or diffusion decoder defines the generation target; \textbf{Upper-U isolated}: generation gradients do not update the upper understanding transformer; \textbf{Train-only aux.}: generation-specific machinery is introduced for understanding training and removed at deployment. \textcolor{orange}{\pmark} denotes partial or architecture-dependent support.
    \end{minipage}
\end{table}

Recent unified multimodal models (UMMs) attempt to close this gap by jointly training understanding and generation within a single framework, thereby introducing visual synthesis objectives that force the model to engage more deeply with image structure. 
One line follows a fully autoregressive formulation: Chameleon \cite{team2024chameleon} and Emu3 \cite{wangMultimodalLearningNexttoken2026} cast both understanding and generation into next-token prediction over a shared discrete visual token space, offering a clean interface for mixed-modality sequence modeling. 
Another line adopts AR-diffusion hybrid formulations, combining language modeling for text with diffusion or flow-based objectives for visual generation. Representative works include Transfusion \cite{zhou2025transfusion}, Show-o \cite{xie2025showo,xie2026showo}, and BLIP3-o \cite{chen2025blip3}. 
Within this family, several works further explore architectural decoupling to mitigate the well-known conflict between generation and understanding objectives: Janus \cite{chen2025janus} decouples visual encoding pathways while sharing a single LLM backbone; MetaMorph \cite{tong2025metamorph} introduces Visual-Predictive Instruction Tuning and observes that improving visual understanding implicitly benefits generation; LMFusion \cite{shi2026lmfusion} adds modality-specific transformer layers to preserve pretrained language capabilities during generation training; and BAGEL \cite{deng2025emerging} scales on trillions of interleaved tokens and exhibits compositional generation and reasoning.

Most closely related to our goal, ROSS~\cite{wang2025reconstructive} reconstructs the input image through a denoising objective over continuous appearance features, whereas ASVR~\cite{wang2026autoregressive} autoregressively reconstructs discrete semantic visual tokens. 
UniHetero~\cite{chen2025unihetero} further shows at large data scale that semantic generation---including autoregression on the LLM input embedding---can improve understanding, while pixel-level generation gradients can be detrimental. UniMRG~\cite{su2026generation} post-trains existing UMMs to generate RGB reconstructions, depth maps, and segmentation maps from an input image. These results strengthen the broad premise that visual-side supervision can improve understanding; accordingly, GAS's distinction is the combination of instruction-conditioned \emph{output-image} prediction across diverse generation tasks, direct supervision in the understanding input space, asymmetric upper-layer isolation, and a generation branch used only during training. 

Appendix Table~\ref{tab:aligned_auxiliary_controls} further compares adapted ROSS-style and ASVR-style objectives under the same backbone, data mixture, and optimization setting. These controls test how their reconstruction objectives transfer to our generation-augmented setting rather than replacing the original papers' reported results.

However, despite these architectural innovations, existing UMMs predominantly pursue generation quality or explore emergent capabilities from scale, while treating the understanding branch as a secondary beneficiary rather than the primary optimization target. Moreover, the introduction of generation objectives frequently degrades understanding performance due to conflicting gradient signals, a tension that current methods address by either accepting the tradeoff or increasing model capacity~\cite{team2026longcat}. Currently, there exists no systematic investigation into methods for integrating visual generation supervision to preserve and further advance models’ semantic understanding capabilities.
\section{Conclusion and Future Work}
\label{sec:conclusion}

We present GAS, a generation-guided training framework that improves aggregate multimodal understanding by using visual generation purely as training-time auxiliary supervision. Concretely, NEP provides a generation target aligned with the LLM input space, while the removable MoT branch mitigates direct optimization conflict and preserves zero inference overhead. Across diverse understanding benchmarks, GAS delivers consistent gains, while representation analyses associate the gains with better visual retention, more focused attention, and task-dependent transfer shaped by generation-task correlation.

These results suggest that visual generation is most effective not as an inference-time capability to preserve, but as a carefully structured supervisory signal for building stronger understanding models. We believe this work establishes generation-guided training as a principled and practical paradigm for advancing multimodal understanding. Looking forward, several promising directions emerge: scaling the framework to longer-context scenarios, extending the NEP paradigm to video-level temporal generation tasks, exploring adaptive task weighting strategies that dynamically balance generation categories during training, and investigating whether the insights on generation-understanding synergy generalize to other modalities such as audio and 3D perception.

\bibliographystyle{unsrtnat}
\bibliography{ref}

@misc{openaiGPT5,
    title = {gpt-5-system-card},
    author = {OpenAI},
    year = 2025
}

@misc{gemini3flash,
    title = {Gemini 3 Flash: frontier intelligence built for speed},
    author = {Google},
    year = 2025
}

@misc{kakaobrain2022coyo-700m,
  title         = {COYO-700M: Image-Text Pair Dataset},
  author        = {Byeon, Minwoo and Park, Beomhee and Kim, Haecheon and Lee, Sungjun and Baek, Woonhyuk and Kim, Saehoon},
  year          = {2022},
  howpublished  = {\url{https://github.com/kakaobrain/coyo-dataset}},
}

@inproceedings{wang2025reconstructive,
  title={Reconstructive visual instruction tuning},
  author={Wang, Haochen and Zheng, Anlin and Zhao, Yucheng and Wang, Tiancai and Ge, Zheng and Zhang, Xiangyu and Zhang, Zhaoxiang},
  booktitle={International Conference on Learning Representations},
  volume={2025},
  pages={14374--14399},
  year={2025}
}

@inproceedings{wang2026autoregressive,
  title={Autoregressive semantic visual reconstruction helps vlms understand better},
  author={Wang, Dianyi and Song, Wei and Wang, Yikun and Wang, Siyuan and Yu, Kaicheng and Wei, Zhongyu and Wang, Jiaqi},
  booktitle={Findings of the Association for Computational Linguistics: ACL 2026},
  pages={38101--38115},
  year={2026}
}

@inproceedings{
su2026generation,
title={Generation Enhances Understanding in Unified Multimodal Models via Multi-Representation Generation},
author={Zihan Su and Hongyang Wei and Kangrui Cen and Yong Wang and Guanhua Chen and Chun Yuan and Xiangxiang Chu},
booktitle={Forty-third International Conference on Machine Learning},
year={2026},
url={https://openreview.net/forum?id=uYaNE2Cys2}
}

@article{li2024llava,
  title={Llava-onevision: Easy visual task transfer},
  author={Li, Bo and Zhang, Yuanhan and Guo, Dong and Zhang, Renrui and Li, Feng and Zhang, Hao and Zhang, Kaichen and Zhang, Peiyuan and Li, Yanwei and Liu, Ziwei and others},
  journal={arXiv preprint arXiv:2408.03326},
  year={2024}
}

@misc{liu2025videoxlproreconstructivetokencompression,
      title={Video-XL-Pro: Reconstructive Token Compression for Extremely Long Video Understanding}, 
      author={Xiangrui Liu and Yan Shu and Zheng Liu and Ao Li and Yang Tian and Bo Zhao},
      year={2025},
      eprint={2503.18478},
      archivePrefix={arXiv},
      primaryClass={cs.CV},
      url={https://arxiv.org/abs/2503.18478}, 
}

@misc{baiQwen25VLTechnicalReport2025,
  title = {Qwen2.5-{{VL Technical Report}}},
  author = {Bai, Shuai and Chen, Keqin and Liu, Xuejing and Wang, Jialin and Ge, Wenbin and Song, Sibo and Dang, Kai and Wang, Peng and Wang, Shijie and Tang, Jun and Zhong, Humen and Zhu, Yuanzhi and Yang, Mingkun and Li, Zhaohai and Wan, Jianqiang and Wang, Pengfei and Ding, Wei and Fu, Zheren and Xu, Yiheng and Ye, Jiabo and Zhang, Xi and Xie, Tianbao and Cheng, Zesen and Zhang, Hang and Yang, Zhibo and Xu, Haiyang and Lin, Junyang},
  year = {2025},
  month = feb,
  number = {arXiv:2502.13923},
  eprint = {2502.13923},
  primaryclass = {cs},
  publisher = {arXiv},
  archiveprefix = {arXiv}
}

@article{kingma2014adam,
  title={Adam: A method for stochastic optimization},
  author={Kingma, Diederik P and Ba, Jimmy},
  journal={arXiv preprint arXiv:1412.6980},
  year={2014}
}

@article{zhu2025internvl3,
  title={Internvl3: Exploring advanced training and test-time recipes for open-source multimodal models},
  author={Zhu, Jinguo and Wang, Weiyun and Chen, Zhe and Liu, Zhaoyang and Ye, Shenglong and Gu, Lixin and Tian, Hao and Duan, Yuchen and Su, Weijie and Shao, Jie and others},
  journal={arXiv preprint arXiv:2504.10479},
  year={2025}
}

@inproceedings{liu2024improved,
  title={Improved baselines with visual instruction tuning},
  author={Liu, Haotian and Li, Chunyuan and Li, Yuheng and Lee, Yong Jae},
  booktitle={Proceedings of the IEEE/CVF conference on computer vision and pattern recognition},
  pages={26296--26306},
  year={2024}
}

@misc{baiQwen3VLTechnicalReport2025,
    title = {Qwen3-{VL} {Technical} {Report}},
    url = {http://arxiv.org/abs/2511.21631},
    doi = {10.48550/arXiv.2511.21631},
    urldate = {2025-11-27},
    publisher = {arXiv},
    author = {Bai, Shuai and Cai, Yuxuan and Chen, Ruizhe and Chen, Keqin and Chen, Xionghui and Cheng, Zesen and Deng, Lianghao and Ding, Wei and Gao, Chang and Ge, Chunjiang and Ge, Wenbin and Guo, Zhifang and Huang, Qidong and Huang, Jie and Huang, Fei and Hui, Binyuan and Jiang, Shutong and Li, Zhaohai and Li, Mingsheng and Li, Mei and Li, Kaixin and Lin, Zicheng and Lin, Junyang and Liu, Xuejing and Liu, Jiawei and Liu, Chenglong and Liu, Yang and Liu, Dayiheng and Liu, Shixuan and Lu, Dunjie and Luo, Ruilin and Lv, Chenxu and Men, Rui and Meng, Lingchen and Ren, Xuancheng and Ren, Xingzhang and Song, Sibo and Sun, Yuchong and Tang, Jun and Tu, Jianhong and Wan, Jianqiang and Wang, Peng and Wang, Pengfei and Wang, Qiuyue and Wang, Yuxuan and Xie, Tianbao and Xu, Yiheng and Xu, Haiyang and Xu, Jin and Yang, Zhibo and Yang, Mingkun and Yang, Jianxin and Yang, An and Yu, Bowen and Zhang, Fei and Zhang, Hang and Zhang, Xi and Zheng, Bo and Zhong, Humen and Zhou, Jingren and Zhou, Fan and Zhou, Jing and Zhu, Yuanzhi and Zhu, Ke},
    month = nov,
    year = {2025},
    note = {arXiv:2511.21631 [cs]},
}

@misc{liuVisualInstructionTuning2023,
    title = {Visual {Instruction} {Tuning}},
    url = {http://arxiv.org/abs/2304.08485},
    language = {en},
    urldate = {2024-09-18},
    publisher = {arXiv},
    author = {Liu, Haotian and Li, Chunyuan and Wu, Qingyang and Lee, Yong Jae},
    month = dec,
    year = {2023},
    note = {arXiv:2304.08485 [cs]},
}

@article{li2024llavanext,
  title={Llava-next-interleave: Tackling multi-image, video, and 3d in large multimodal models},
  author={Li, Feng and Zhang, Renrui and Zhang, Hao and Zhang, Yuanhan and Li, Bo and Li, Wei and Ma, Zejun and Li, Chunyuan},
  journal={arXiv preprint arXiv:2407.07895},
  year={2024}
}

@article{alayrac2022flamingo,
  title={Flamingo: a visual language model for few-shot learning},
  author={Alayrac, Jean-Baptiste and Donahue, Jeff and Luc, Pauline and Miech, Antoine and Barr, Iain and Hasson, Yana and Lenc, Karel and Mensch, Arthur and Millican, Katherine and Reynolds, Malcolm and others},
  journal={Advances in neural information processing systems},
  volume={35},
  pages={23716--23736},
  year={2022}
}

@article{dai2023instructblip,
  title={Instructblip: Towards general-purpose vision-language models with instruction tuning},
  author={Dai, Wenliang and Li, Junnan and Li, Dongxu and Tiong, Anthony and Zhao, Junqi and Wang, Weisheng and Li, Boyang and Fung, Pascale N and Hoi, Steven},
  journal={Advances in neural information processing systems},
  volume={36},
  pages={49250--49267},
  year={2023}
}

@inproceedings{radford2021learning,
  title={Learning transferable visual models from natural language supervision},
  author={Radford, Alec and Kim, Jong Wook and Hallacy, Chris and Ramesh, Aditya and Goh, Gabriel and Agarwal, Sandhini and Sastry, Girish and Askell, Amanda and Mishkin, Pamela and Clark, Jack and others},
  booktitle={International conference on machine learning},
  pages={8748--8763},
  year={2021},
  organization={PmLR}
}

@article{tschannen2025siglip,
  title={Siglip 2: Multilingual vision-language encoders with improved semantic understanding, localization, and dense features},
  author={Tschannen, Michael and Gritsenko, Alexey and Wang, Xiao and Naeem, Muhammad Ferjad and Alabdulmohsin, Ibrahim and Parthasarathy, Nikhil and Evans, Talfan and Beyer, Lucas and Xia, Ye and Mustafa, Basil and others},
  journal={arXiv preprint arXiv:2502.14786},
  year={2025}
}

@misc{IntroducingGemini202024,
    title = {Introducing {Gemini} 2.0: our new {AI} model for the agentic era},
    shorttitle = {Introducing {Gemini} 2.0},
    url = {https://blog.google/technology/google-deepmind/google-gemini-ai-update-december-2024/},
    language = {en-us},
    urldate = {2025-01-26},
    journal = {Google},
    month = dec,
    year = {2024},
}

@inproceedings{shu2025video,
  title={Video-xl: Extra-long vision language model for hour-scale video understanding},
  author={Shu, Yan and Liu, Zheng and Zhang, Peitian and Qin, Minghao and Zhou, Junjie and Liang, Zhengyang and Huang, Tiejun and Zhao, Bo},
  booktitle={Proceedings of the Computer Vision and Pattern Recognition Conference},
  pages={26160--26169},
  year={2025}
}

@inproceedings{
    li2026videochatflash,
    title={VideoChat-Flash: Hierarchical Compression for Long-Context Video Modeling},
    author={Xinhao Li and Yi Wang and Jiashuo Yu and Xiangyu Zeng and Yuhan Zhu and Haian Huang and Jianfei Gao and Kunchang Li and Yinan He and Chenting Wang and Yu Qiao and Yali Wang and Limin Wang},
    booktitle={The Fourteenth International Conference on Learning Representations},
    year={2026},
    url={https://openreview.net/forum?id=MUjdNcfNPv}
}

@inproceedings{
    xu2026visulogic,
    title={VisuLogic: A Benchmark for Evaluating Visual Reasoning in Multi-modal Large Language Models},
    author={Weiye Xu and Jiahao Wang and Weiyun Wang and Zhe Chen and Wengang Zhou and Aijun Yang and Lewei Lu and Houqiang Li and Xiaohua Wang and Xizhou Zhu and Wenhai Wang and Jifeng Dai and Jinguo Zhu},
    booktitle={The Fourteenth International Conference on Learning Representations},
    year={2026},
    url={https://openreview.net/forum?id=mXuzDDVXxi}
}

@inproceedings{
    wang2026vgr,
    title={{VGR}: Visual Grounded Reasoning},
    author={Jiacong Wang and Zijian Kang and Haochen Wang and LiangXiao and Ya Wang and Jiawen Li and Bohong Wu and Ran Jiao and Haiyong Jiang and ChaoFeng and Jun Xiao},
    booktitle={The Fourteenth International Conference on Learning Representations},
    year={2026},
    url={https://openreview.net/forum?id=kDhAiaGzrn}
}

@article{ke2025explain,
  title={Explain before you answer: A survey on compositional visual reasoning},
  author={Ke, Fucai and Hsu, Joy and Cai, Zhixi and Ma, Zixian and Zheng, Xin and Wu, Xindi and Huang, Sukai and Wang, Weiqing and Haghighi, Pari Delir and Haffari, Gholamreza and others},
  journal={arXiv preprint arXiv:2508.17298},
  year={2025}
}

@article{team2024chameleon,
  title={Chameleon: Mixed-modal early-fusion foundation models},
  author={Team, Chameleon},
  journal={arXiv preprint arXiv:2405.09818},
  year={2024}
}

@article{wangMultimodalLearningNexttoken2026,
  title = {Multimodal Learning with Next-Token Prediction for Large Multimodal Models},
  author = {Wang, Xinlong and Cui, Yufeng and Wang, Jinsheng and Zhang, Fan and Wang, Yueze and Zhang, Xiaosong and Luo, Zhengxiong and Sun, Quan and Li, Zhen and Wang, Yuqi and Yu, Qiying and Zhao, Yingli and Ao, Yulong and Min, Xuebin and Men, Chunlei and Wu, Boya and Zhao, Bo and Zhang, Bowen and Wang, Liangdong and Liu, Guang and He, Zheqi and Yang, Xi and Liu, Jingjing and Lin, Yonghua and Wang, Zhongyuan and Huang, Tiejun},
  year = 2026,
  month = feb,
  journal = {Nature},
  volume = {650},
  number = {8101},
  pages = {327--333},
  issn = {1476-4687},
  doi = {10.1038/s41586-025-10041-x}
}

@inproceedings{
    zhou2025transfusion,
    title={Transfusion: Predict the Next Token and Diffuse Images with One Multi-Modal Model},
    author={Chunting Zhou and LILI YU and Arun Babu and Kushal Tirumala and Michihiro Yasunaga and Leonid Shamis and Jacob Kahn and Xuezhe Ma and Luke Zettlemoyer and Omer Levy},
    booktitle={The Thirteenth International Conference on Learning Representations},
    year={2025},
    url={https://openreview.net/forum?id=SI2hI0frk6}
}

@inproceedings{
    xie2025showo,
    title={Show-o: One Single Transformer to Unify Multimodal Understanding and Generation},
    author={Jinheng Xie and Weijia Mao and Zechen Bai and David Junhao Zhang and Weihao Wang and Kevin Qinghong Lin and Yuchao Gu and Zhijie Chen and Zhenheng Yang and Mike Zheng Shou},
    booktitle={The Thirteenth International Conference on Learning Representations},
    year={2025},
    url={https://openreview.net/forum?id=o6Ynz6OIQ6}
}

@inproceedings{
    xie2026showo,
    title={Show-o2: Improved Native Unified Multimodal Models},
    author={Jinheng Xie and Zhenheng Yang and Mike Zheng Shou},
    booktitle={The Thirty-ninth Annual Conference on Neural Information Processing Systems},
    year={2026},
    url={https://openreview.net/forum?id=7VMg7Jb7AL}
}

@article{chen2025blip3,
  title={Blip3-o: A family of fully open unified multimodal models-architecture, training and dataset},
  author={Chen, Jiuhai and Xu, Zhiyang and Pan, Xichen and Hu, Yushi and Qin, Can and Goldstein, Tom and Huang, Lifu and Zhou, Tianyi and Xie, Saining and Savarese, Silvio and others},
  journal={arXiv preprint arXiv:2505.09568},
  year={2025}
}

@article{chen2025janus,
  title={Janus-pro: Unified multimodal understanding and generation with data and model scaling},
  author={Chen, Xiaokang and Wu, Zhiyu and Liu, Xingchao and Pan, Zizheng and Liu, Wen and Xie, Zhenda and Yu, Xingkai and Ruan, Chong},
  journal={arXiv preprint arXiv:2501.17811},
  year={2025}
}

@inproceedings{
    shi2026lmfusion,
    title={{LMF}usion: Adapting Pretrained Language Models for Multimodal Generation},
    author={Weijia Shi and Xiaochuang Han and Chunting Zhou and Weixin Liang and Xi Victoria Lin and Luke Zettlemoyer and LILI YU},
    booktitle={The Thirty-ninth Annual Conference on Neural Information Processing Systems},
    year={2026},
    url={https://openreview.net/forum?id=Kc1WTxZbrP}
}

@article{deng2025emerging,
  title={Emerging properties in unified multimodal pretraining},
  author={Deng, Chaorui and Zhu, Deyao and Li, Kunchang and Gou, Chenhui and Li, Feng and Wang, Zeyu and Zhong, Shu and Yu, Weihao and Nie, Xiaonan and Song, Ziang and others},
  journal={arXiv preprint arXiv:2505.14683},
  year={2025}
}

@article{fu2026lance,
  title={Lance: Unified Multimodal Modeling by Multi-Task Synergy},
  author={Fu, Fengyi and Huang, Mengqi and Wu, Shaojin and Jiang, Yunsheng and Huo, Yufei and Li, Hao and Song, Yinghang and Ding, Fei and Guo, Jianzhu and He, Qian and Fu, Zheren and Mao, Zhendong and Zhang, Yongdong},
  journal={arXiv preprint arXiv:2605.18678},
  year={2026}
}

@article{shen2025mammothmoda2,
  title={MammothModa2: A Unified AR-Diffusion Framework for Multimodal Understanding and Generation},
  author={Shen, Tao and Wan, Xin and Chen, Taicai and Zhang, Rui and Pan, Junwen and Lu, Dawei and Lei, Fanding and Lu, Zhilin and Yang, Yunfei and Cheng, Chen and others},
  journal={arXiv preprint arXiv:2511.18262},
  year={2025}
}

@article{team2026longcat,
  title={Longcat-next: Lexicalizing modalities as discrete tokens},
  author={Team, Meituan LongCat and Xiao, Bin and Wang, Chao and Li, Chengjiang and Zhang, Chi and Peng, Chong and Yu, Hang and Yang, Hao and Yan, Haonan and Sun, Haoze and others},
  journal={arXiv preprint arXiv:2603.27538},
  year={2026}
}

@inproceedings{shiri2024empirical,
  title={An empirical analysis on spatial reasoning capabilities of large multimodal models},
  author={Shiri, Fatemeh and Guo, Xiao-Yu and Far, Mona Golestan and Yu, Xin and Haf, Reza and Li, Yuan-Fang},
  booktitle={Proceedings of the 2024 Conference on Empirical Methods in Natural Language Processing},
  pages={21440--21455},
  year={2024}
}

@article{shao2024visual,
  title={Visual cot: Advancing multi-modal language models with a comprehensive dataset and benchmark for chain-of-thought reasoning},
  author={Shao, Hao and Qian, Shengju and Xiao, Han and Song, Guanglu and Zong, Zhuofan and Wang, Letian and Liu, Yu and Li, Hongsheng},
  journal={Advances in Neural Information Processing Systems},
  volume={37},
  pages={8612--8642},
  year={2024}
}

@article{zhang2024visually,
  title={Why are visually-grounded language models bad at image classification?},
  author={Zhang, Yuhui and Unell, Alyssa and Wang, Xiaohan and Ghosh, Dhruba and Su, Yuchang and Schmidt, Ludwig and Yeung-Levy, Serena},
  journal={Advances in Neural Information Processing Systems},
  volume={37},
  pages={51727--51753},
  year={2024}
}

@inproceedings{ganz2024question,
  title={Question aware vision transformer for multimodal reasoning},
  author={Ganz, Roy and Kittenplon, Yair and Aberdam, Aviad and Ben Avraham, Elad and Nuriel, Oren and Mazor, Shai and Litman, Ron},
  booktitle={Proceedings of the IEEE/CVF conference on computer vision and pattern recognition},
  pages={13861--13871},
  year={2024}
}

@inproceedings{wu2025janus,
  title={Janus: Decoupling visual encoding for unified multimodal understanding and generation},
  author={Wu, Chengyue and Chen, Xiaokang and Wu, Zhiyu and Ma, Yiyang and Liu, Xingchao and Pan, Zizheng and Liu, Wen and Xie, Zhenda and Yu, Xingkai and Ruan, Chong and others},
  booktitle={Proceedings of the Computer Vision and Pattern Recognition Conference},
  pages={12966--12977},
  year={2025}
}

@article{yang2026mmada,
  title={Mmada: Multimodal large diffusion language models},
  author={Yang, Ling and Tian, Ye and Li, Bowen and Zhang, Xinchen and Shen, Ke and Tong, Yunhai and Wang, Mengdi},
  journal={Advances in Neural Information Processing Systems},
  volume={38},
  pages={138867--138907},
  year={2026}
}

@article{xu2025next,
  title={Next-Embedding Prediction Makes Strong Vision Learners},
  author={Xu, Sihan and Ma, Ziqiao and Chai, Wenhao and Chen, Xuweiyi and Jin, Weiyang and Chai, Joyce and Xie, Saining and Yu, Stella X},
  journal={arXiv preprint arXiv:2512.16922},
  year={2025}
}

@article{
  liang2025mixtureoftransformers,
  title={Mixture-of-Transformers: A Sparse and Scalable Architecture for Multi-Modal Foundation Models},
  author={Weixin Liang and LILI YU and Liang Luo and Srini Iyer and Ning Dong and Chunting Zhou and Gargi Ghosh and Mike Lewis and Wen-tau Yih and Luke Zettlemoyer and Xi Victoria Lin},
  journal={Transactions on Machine Learning Research},
  issn={2835-8856},
  year={2025},
  url={https://openreview.net/forum?id=Nu6N69i8SB},
  note={}
}

@article{fu2026mme,
  title={Mme: A comprehensive evaluation benchmark for multimodal large language models},
  author={Fu, Chaoyou and Chen, Peixian and Shen, Yunhang and Qin, Yulei and Zhang, Mengdan and Lin, Xu and Yang, Jinrui and Zheng, Xiawu and Li, Ke and Sun, Xing and others},
  journal={Advances in Neural Information Processing Systems},
  volume={38},
  year={2026}
}

@inproceedings{yue2024mmmu,
  title={Mmmu: A massive multi-discipline multimodal understanding and reasoning benchmark for expert agi},
  author={Yue, Xiang and Ni, Yuansheng and Zhang, Kai and Zheng, Tianyu and Liu, Ruoqi and Zhang, Ge and Stevens, Samuel and Jiang, Dongfu and Ren, Weiming and Sun, Yuxuan and others},
  booktitle={Proceedings of the IEEE/CVF conference on computer vision and pattern recognition},
  pages={9556--9567},
  year={2024}
}

@inproceedings{fu2024blink,
  title={Blink: Multimodal large language models can see but not perceive},
  author={Fu, Xingyu and Hu, Yushi and Li, Bangzheng and Feng, Yu and Wang, Haoyu and Lin, Xudong and Roth, Dan and Smith, Noah A and Ma, Wei-Chiu and Krishna, Ranjay},
  booktitle={European Conference on Computer Vision},
  pages={148--166},
  year={2024},
  organization={Springer}
}

@article{wang2024charxiv,
  title={Charxiv: Charting gaps in realistic chart understanding in multimodal llms},
  author={Wang, Zirui and Xia, Mengzhou and He, Luxi and Chen, Howard and Liu, Yitao and Zhu, Richard and Liang, Kaiqu and Wu, Xindi and Liu, Haotian and Malladi, Sadhika and others},
  journal={Advances in Neural Information Processing Systems},
  volume={37},
  pages={113569--113697},
  year={2024}
}

@inproceedings{zou2025dynamath,
  title={Dynamath: A dynamic visual benchmark for evaluating mathematical reasoning robustness of vision language models},
  author={Zou, Chengke and Guo, Xingang and Yang, Rui and Zhang, Junyu and Hu, Bin and Zhang, Huan},
  booktitle={International Conference on Learning Representations},
  volume={2025},
  pages={48337--48383},
  year={2025}
}

@article{wang2024measuring,
  title={Measuring multimodal mathematical reasoning with math-vision dataset},
  author={Wang, Ke and Pan, Junting and Shi, Weikang and Lu, Zimu and Ren, Houxing and Zhou, Aojun and Zhan, Mingjie and Li, Hongsheng},
  journal={Advances in Neural Information Processing Systems},
  volume={37},
  pages={95095--95169},
  year={2024}
}

@inproceedings{lu2024mathvista,
  title={Mathvista: Evaluating mathematical reasoning of foundation models in visual contexts},
  author={Lu, Pan and Bansal, Hritik and Xia, Tony and Liu, Jiacheng and Li, Chunyuan and Hajishirzi, Hannaneh and Cheng, Hao and Chang, Kai-Wei and Galley, Michel and Gao, Jianfeng},
  booktitle={International Conference on Learning Representations},
  volume={2024},
  pages={23439--23554},
  year={2024}
}

@article{xiao2024logicvista,
  title={Logicvista: Multimodal llm logical reasoning benchmark in visual contexts},
  author={Xiao, Yijia and Sun, Edward and Liu, Tianyu and Wang, Wei},
  journal={arXiv preprint arXiv:2407.04973},
  year={2024}
}

@article{xu2025visulogic,
  title={Visulogic: A benchmark for evaluating visual reasoning in multi-modal large language models},
  author={Xu, Weiye and Wang, Jiahao and Wang, Weiyun and Chen, Zhe and Zhou, Wengang and Yang, Aijun and Lu, Lewei and Li, Houqiang and Wang, Xiaohua and Zhu, Xizhou and others},
  journal={arXiv preprint arXiv:2504.15279},
  year={2025}
}

@article{beyer2024paligemma,
      title={{PaliGemma: A versatile 3B VLM for transfer}},
      author={Lucas Beyer and Andreas Steiner and André Susano Pinto and Alexander Kolesnikov and Xiao Wang and Daniel Salz and Maxim Neumann and Ibrahim Alabdulmohsin and Michael Tschannen and Emanuele Bugliarello and Thomas Unterthiner and Daniel Keysers and Skanda Koppula and Fangyu Liu and Adam Grycner and Alexey Gritsenko and Neil Houlsby and Manoj Kumar and Keran Rong and Julian Eisenschlos and Rishabh Kabra and Matthias Bauer and Matko Bošnjak and Xi Chen and Matthias Minderer and Paul Voigtlaender and Ioana Bica and Ivana Balazevic and Joan Puigcerver and Pinelopi Papalampidi and Olivier Henaff and Xi Xiong and Radu Soricut and Jeremiah Harmsen and Xiaohua Zhai},
      year={2024},
      journal={arXiv preprint arXiv:2407.07726}
}

@article{paiss2023countclip,
      title={{Teaching CLIP to Count to Ten}},
      author={Paiss, Roni and Ephrat, Ariel and Tov, Omer and Zada, Shiran and Mosseri, Inbar and Irani, Michal and Dekel, Tali},
      year={2023},
      journal={arXiv preprint arXiv:2302.12066}
}

@article{tong2024cambrian,
  title={Cambrian-1: A fully open, vision-centric exploration of multimodal llms},
  author={Tong, Shengbang and Brown, Ellis and Wu, Penghao and Woo, Sanghyun and Middepogu, Manoj and Akula, Sai C and Yang, Jihan and Yang, Shusheng and Iyer, Adithya and Pan, Xichen and others},
  journal={Advances in Neural Information Processing Systems},
  volume={37},
  pages={87310--87356},
  year={2024}
}

@inproceedings{fu2025video,
  title={Video-mme: The first-ever comprehensive evaluation benchmark of multi-modal llms in video analysis},
  author={Fu, Chaoyou and Dai, Yuhan and Luo, Yongdong and Li, Lei and Ren, Shuhuai and Zhang, Renrui and Wang, Zihan and Zhou, Chenyu and Shen, Yunhang and Zhang, Mengdan and others},
  booktitle={Proceedings of the IEEE/CVF conference on computer vision and pattern recognition},
  pages={24108--24118},
  year={2025}
}

@inproceedings{li2024mvbench,
  title={Mvbench: A comprehensive multi-modal video understanding benchmark},
  author={Li, Kunchang and Wang, Yali and He, Yinan and Li, Yizhuo and Wang, Yi and Liu, Yi and Wang, Zun and Xu, Jilan and Chen, Guo and Luo, Ping and others},
  booktitle={Proceedings of the IEEE/CVF Conference on Computer Vision and Pattern Recognition},
  pages={22195--22206},
  year={2024}
}

@article{yang2025qwen3,
  title={Qwen3 technical report},
  author={Yang, An and Li, Anfeng and Yang, Baosong and Zhang, Beichen and Hui, Binyuan and Zheng, Bo and Yu, Bowen and Gao, Chang and Huang, Chengen and Lv, Chenxu and others},
  journal={arXiv preprint arXiv:2505.09388},
  year={2025}
}

@inproceedings{deng2009imagenet,
  title={Imagenet: A large-scale hierarchical image database},
  author={Deng, Jia and Dong, Wei and Socher, Richard and Li, Li-Jia and Li, Kai and Fei-Fei, Li},
  booktitle={2009 IEEE conference on computer vision and pattern recognition},
  pages={248--255},
  year={2009},
  organization={Ieee}
}

@inproceedings{mao2016generation,
  title={Generation and comprehension of unambiguous object descriptions},
  author={Mao, Junhua and Huang, Jonathan and Toshev, Alexander and Camburu, Oana and Yuille, Alan L and Murphy, Kevin},
  booktitle={Proceedings of the IEEE conference on computer vision and pattern recognition},
  pages={11--20},
  year={2016}
}

@inproceedings{kazemzadeh2014referitgame,
  title={Referitgame: Referring to objects in photographs of natural scenes},
  author={Kazemzadeh, Sahar and Ordonez, Vicente and Matten, Mark and Berg, Tamara},
  booktitle={Proceedings of the 2014 conference on empirical methods in natural language processing (EMNLP)},
  pages={787--798},
  year={2014}
}

@article{lin2025zebralogic,
  title={Zebralogic: On the scaling limits of llms for logical reasoning},
  author={Lin, Bill Yuchen and Bras, Ronan Le and Richardson, Kyle and Sabharwal, Ashish and Poovendran, Radha and Clark, Peter and Choi, Yejin},
  journal={arXiv preprint arXiv:2502.01100},
  year={2025}
}

@inproceedings{gema2025we,
  title={Are we done with mmlu?},
  author={Gema, Aryo Pradipta and Leang, Joshua Ong Jun and Hong, Giwon and Devoto, Alessio and Mancino, Alberto Carlo Maria and Saxena, Rohit and He, Xuanli and Zhao, Yu and Du, Xiaotang and Madani, Mohammad Reza Ghasemi and others},
  booktitle={Proceedings of the 2025 Conference of the Nations of the Americas Chapter of the Association for Computational Linguistics: Human Language Technologies (Volume 1: Long Papers)},
  pages={5069--5096},
  year={2025}
}

@inproceedings{tong2025metamorph,
  title={Metamorph: Multimodal understanding and generation via instruction tuning},
  author={Tong, Shengbang and Fan, David and Li, Jiachen and Xiong, Yunyang and Chen, Xinlei and Sinha, Koustuv and Rabbat, Michael and LeCun, Yann and Xie, Saining and Liu, Zhuang},
  booktitle={Proceedings of the IEEE/CVF International Conference on Computer Vision},
  pages={17001--17012},
  year={2025}
}

@article{zhang2026cheers,
  title={Cheers: Decoupling Patch Details from Semantic Representations Enables Unified Multimodal Comprehension and Generation},
  author={Zhang, Yichen and Peng, Da and Guo, Zonghao and Zhang, Zijian and Yang, Xuesong and Sun, Tong and Sun, Shichu and Zhang, Yidan and Li, Yanghao and Zhao, Haiyan and others},
  journal={arXiv preprint arXiv:2603.12793},
  year={2026}
}

@article{chen2025unihetero,
  title={UniHetero: Could Generation Enhance Understanding for Vision-Language-Model at Large Data Scale?},
  author={Chen, Fengjiao and Jing, Minhao and Lu, Weitao and Feng, Yan and Li, Xiaoyu and Cao, Xuezhi},
  journal={arXiv preprint arXiv:2512.23512},
  year={2025}
}

@article{li2025unifork,
  title={Unifork: Exploring modality alignment for unified multimodal understanding and generation},
  author={Li, Teng and Lu, Quanfeng and Zhao, Lirui and Li, Hao and Zhu, Xizhou and Qiao, Yu and Zhang, Jun and Shao, Wenqi},
  journal={arXiv preprint arXiv:2506.17202},
  year={2025}
}

@article{hao2025uni,
  title={Uni-X: Mitigating Modality Conflict with a Two-End-Separated Architecture for Unified Multimodal Models},
  author={Hao, Jitai and Liu, Hao and Xiao, Xinyan and Huang, Qiang and Yu, Jun},
  journal={arXiv preprint arXiv:2509.24365},
  year={2025}
}

@inproceedings{rajbhandari2020zero,
  title={Zero: Memory optimizations toward training trillion parameter models},
  author={Rajbhandari, Samyam and Rasley, Jeff and Ruwase, Olatunji and He, Yuxiong},
  booktitle={SC20: international conference for high performance computing, networking, storage and analysis},
  pages={1--16},
  year={2020},
  organization={IEEE}
}

@inproceedings{
    zhang2026thyme,
    title={Thyme: Think Beyond Images},
    author={YiFan Zhang and Xingyu Lu and Shukang Yin and Chaoyou Fu and Wei Chen and Xiao Hu and Bin Wen and Kaiyu Jiang and Changyi Liu and Tianke Zhang and Haonan fan and Kaibing Chen and Jiankang Chen and Haojie Ding and Kaiyu Tang and Zhang Zhang and Liang Wang and Fan Yang and Tingting Gao and Guorui Zhou},
    booktitle={The Fourteenth International Conference on Learning Representations},
    year={2026},
    url={https://openreview.net/forum?id=gCWLkqK45O}
}

\appendix

\section{Additional Controlled Experiments}
\label{app:additional_controls}

\subsection{Repeated-Run Robustness Across Data Scales}
\label{app:robustness}

We repeat GAS three times in the 2.5B-token regime and twice in the 20B-token regime. Table~\ref{tab:robustness} reports the sample mean and sample standard deviation, together with the original and longer-trained understanding-only controls. The GAS mean remains above both single-run controls in Overall at each scale. Variation is larger for individual capabilities---most notably Reasoning at 2.5B and Count\&Spatial at 20B---so these runs support reproducible aggregate transfer.

\begin{table}[htbp]
    \renewcommand{\arraystretch}{0.98}
    \centering
    \setlength{\tabcolsep}{3pt}
    \resizebox{\textwidth}{!}{
    \begin{tabular}{@{}llcccccc@{}}
        \toprule
        \textbf{Regime} & \textbf{Configuration} & \textbf{Runs} & \textbf{Overall} & \textbf{Perception} & \textbf{Reasoning} & \textbf{Count\&Spatial} & \textbf{Video} \\
        \midrule
        2.5B & Baseline & 1 & 47.25 & 56.55 & 31.38 & \textbf{76.92} & 46.60 \\
        2.5B & Longer baseline (2.77B; +11\%) & 1 & 47.73 & 57.63 & \textbf{34.01} & 71.28 & 45.55 \\
        2.5B & GAS & 3 & \textbf{$48.46\!\pm\!0.37$} & \textbf{$58.56\!\pm\!0.46$} & $32.99\!\pm\!1.34$ & $75.83\!\pm\!1.06$ & \textbf{$47.28\!\pm\!0.35$} \\
        \midrule
        20B & Baseline & 1 & 51.95 & 60.12 & 38.88 & 79.62 & 47.10 \\
        20B & Longer baseline (22B; +10\%) & 1 & 52.14 & 60.18 & 38.58 & \textbf{80.90} & 46.95 \\
        20B & GAS & 2 & \textbf{$52.33\!\pm\!0.33$} & \textbf{$61.04\!\pm\!0.45$} & \textbf{$39.05\!\pm\!0.24$} & $80.71\!\pm\!1.54$ & \textbf{$47.35\!\pm\!0.49$} \\
        \bottomrule
    \end{tabular}
    }
    \caption{\textbf{Repeated-run robustness across training scales.} GAS entries are mean $\pm$ sample standard deviation; baseline controls are single runs. Bold marks the best reported value within each regime and capability.}
    \label{tab:robustness}
\end{table}

\subsection{Aligned Comparison with Reconstructive Objectives}
\label{app:aligned_auxiliary_controls}

We reproduce ROSS-style and ASVR-style objectives using the same Qwen3-VL-2B backbone, 2.5B understanding plus 2.5B generation data, and optimization setting as GAS. On generation samples, these adapted controls reconstruct the assistant output image using continuous appearance features and discrete semantic tokens, respectively. As shown in Table~\ref{tab:aligned_auxiliary_controls}, input-reconstruction objectives designed for understanding data do not automatically transfer to this generation-augmented setting, whereas GAS's task-conditioned output prediction gives the strongest aggregate result. These are controlled adaptations for isolating the objective under our setup, rather than reproductions of the original papers' reported training regimes.

\begin{table}[htbp]
    \renewcommand{\arraystretch}{0.98}
    \centering
    \setlength{\tabcolsep}{4pt}
    \begin{tabular}{@{}lccccc@{}}
        \toprule
        \textbf{Method} & \textbf{Overall} & \textbf{Perception} & \textbf{Reasoning} & \textbf{Count\&Spatial} & \textbf{Video} \\
        \midrule
        Understanding-only baseline & 47.25 & 56.55 & 31.38 & \textbf{76.92} & 46.60 \\
        ROSS-style (U+G) & 45.70 & 56.10 & 30.88 & 70.80 & 44.25 \\
        ASVR-style (U+G) & 47.31 & 56.20 & \textbf{33.16} & 73.78 & 45.55 \\
        GAS (U+G) & \textbf{48.25} & \textbf{58.52} & 32.63 & 75.15 & \textbf{47.65} \\
        \bottomrule
    \end{tabular}
    \caption{\textbf{Same-backbone comparison with adapted reconstructive objectives.} All U+G methods use the same backbone, data mixture, and optimization setting.}
    \label{tab:aligned_auxiliary_controls}
\end{table}


\section{More Qualitative Cases of GAS}
\label{app:more_qualitative_cases}

\refstepcounter{qualbox}\label{box:visulogic}
\begin{tcolorbox}[qualbox, title = {Qualitative Case~\thequalbox: VisuLogic~\cite{xu2026visulogic}}]

\vspace{\medskipamount}
{\includegraphics[width=0.4\linewidth]{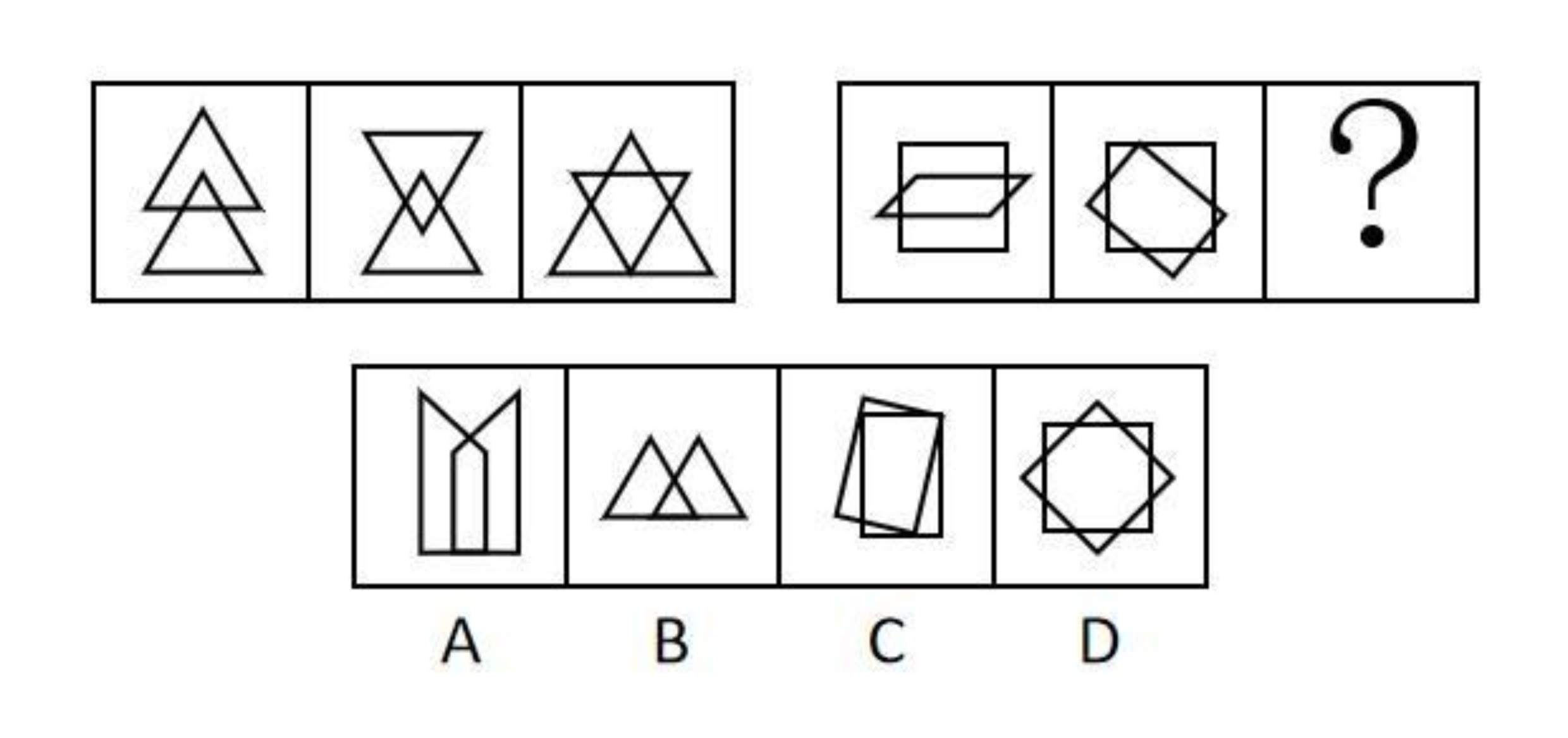}}
\vspace{\medskipamount}

\textbf{Question: }Choose the most appropriate one from the four given options to fill in the question mark so that a certain regularity is presented: A, B, C, D.

\noindent\rule{\linewidth}{0.5pt}

\textbf{Ours 4B: }
{\scriptsize\ttfamily

\textbf{\textcolor{blue}{<think>}}

To solve this problem, we first analyze the shapes and number of lines in each figure, then identify the underlying pattern.

1. **Analyze shape and line count:**

   Observing the figures in the first and second rows, each figure is composed of \textbf{triangles} or \textbf{quadrilaterals} (rhombus-like shapes), with a regular line count.
   
   - Row 1, Fig. 1: two triangles, lines $= 2 \times 3 = 6$;
   
   - Row 1, Fig. 2: two quadrilaterals, lines $= 2 \times 4 = 8$;
   
   - Row 1, Fig. 3: two triangles, lines $= 2 \times 3 = 6$;
   
   - Row 2, Fig. 1: two quadrilaterals, lines $= 2 \times 4 = 8$;
   
   - Row 2, Fig. 2: two triangles, lines $= 2 \times 3 = 6$;
   
   - Row 2, Fig. 3 (the question mark): should follow the same shape/line-count pattern and stacking style.

2. **Identify the pattern:**

   The figures in both rows alternate between \textbf{triangle pairs} and \textbf{quadrilateral pairs}, with line counts of 6 and 8 respectively. All figures share the same two-shape overlapping composition with consistent contour and inner-line connectivity.

3. **Match the options:**

   Option D consists of two overlapping quadrilaterals (rhombus-like), whose line count and composition match the pattern (Row 2 starts with a quadrilateral pair, so the question mark should also be a quadrilateral pair). Option A is a triangle pair with a mismatched shape; Option B is a triangle pair with inconsistent line count; Option C is a quadrilateral but with a different stacking style.

Therefore, the correct answer is $\boxed{D}$.

\textbf{\textcolor{blue}{</think>}}\\

The correct answer is \textbf{\textcolor{tablegreen}{$\boxed{D}$}}.
}

\end{tcolorbox}

\refstepcounter{qualbox}\label{box:dynamath}
\begin{tcolorbox}[qualbox, title = {Qualitative Case~\thequalbox: DynaMath~\cite{zou2025dynamath}}]

\vspace{\medskipamount}
{\includegraphics[width=0.4\linewidth]{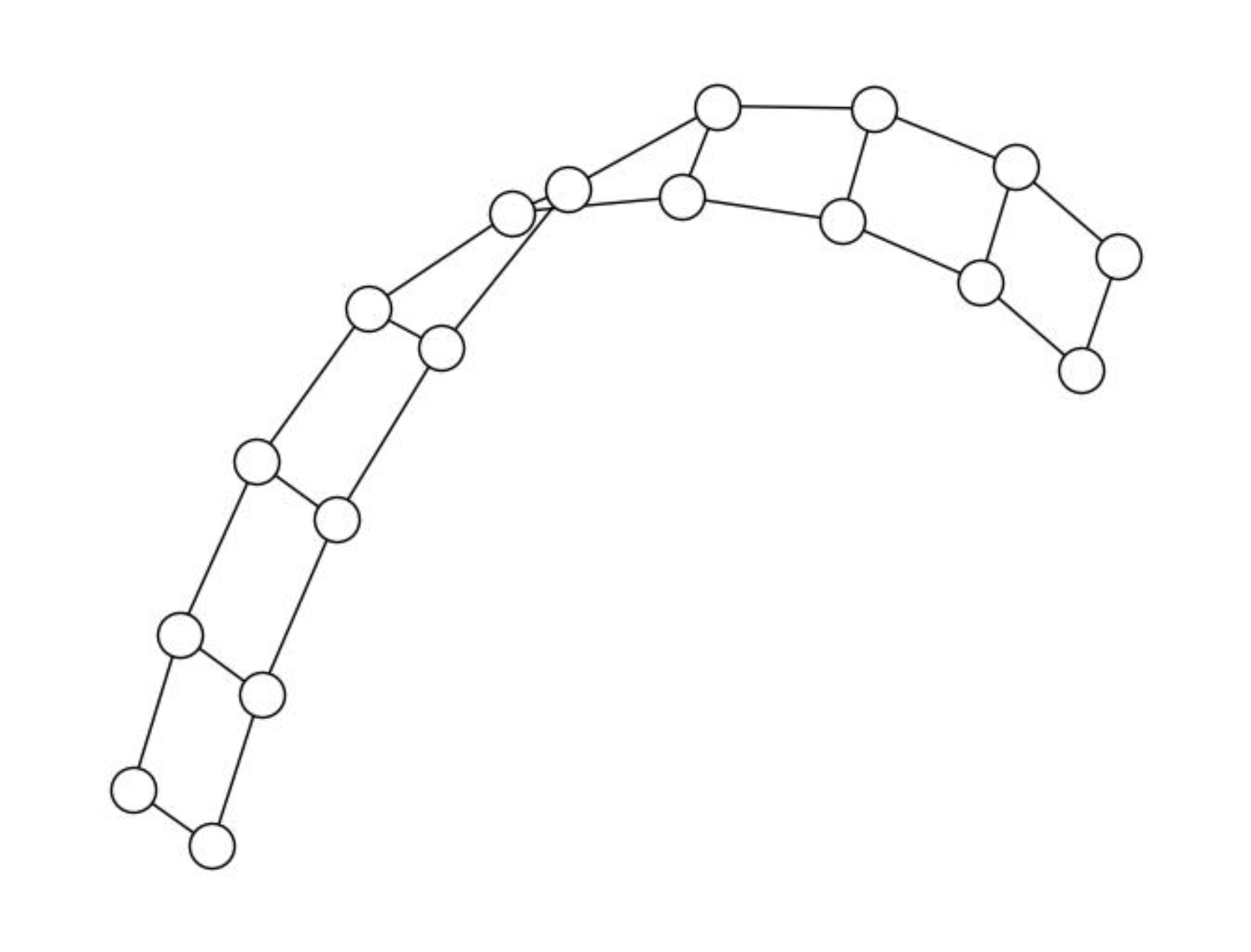}}
\vspace{\medskipamount}

\textbf{Question: }Find the chromatic number of the following graph.

\noindent\rule{\linewidth}{0.5pt}

\textbf{Ours 4B: }
{\scriptsize\ttfamily

\textbf{\textcolor{blue}{<think>}}
To determine the chromatic number of the given graph, we need to find the minimum number of colors required to color the vertices such that no two adjacent vertices share the same color. Let's analyze the graph step by step.
 
1. **Identify the Structure of the Graph:**
  - The graph appears to be a bipartite graph. A bipartite graph is a graph whose vertices can be divided into two disjoint sets such that every edge connects a vertex in one set to a vertex in the other set.
  - In this graph, we can observe that the vertices can be divided into two sets: one set containing the vertices on the left side of the graph and the other set containing the vertices on the right side of the graph.
 
2. **Verify Bipartiteness:**
  - Check if there are any odd-length cycles in the graph. If there are no odd-length cycles, the graph is bipartite.
  - Upon inspection, the graph does not contain any odd-length cycles. All cycles in the graph are even-length cycles.
 
3. **Determine the Chromatic Number:**
  - For a bipartite graph, the chromatic number is always 2. This is because we can color one set of vertices with one color and the other set with a different color, ensuring that no two adjacent vertices share the same color.
 
4. **Conclusion:**
  - Since the graph is bipartite, the chromatic number is 2.

\textbf{\textcolor{blue}{</think>}}\\

The final answer is \textbf{\textcolor{tablegreen}{$\boxed{2}$}}.
}

\end{tcolorbox}

\refstepcounter{qualbox}\label{box:realworldqa}
\begin{tcolorbox}[qualbox, title = {Qualitative Case~\thequalbox: RealWorldQA}]

\vspace{\medskipamount}
{\includegraphics[width=0.4\linewidth]{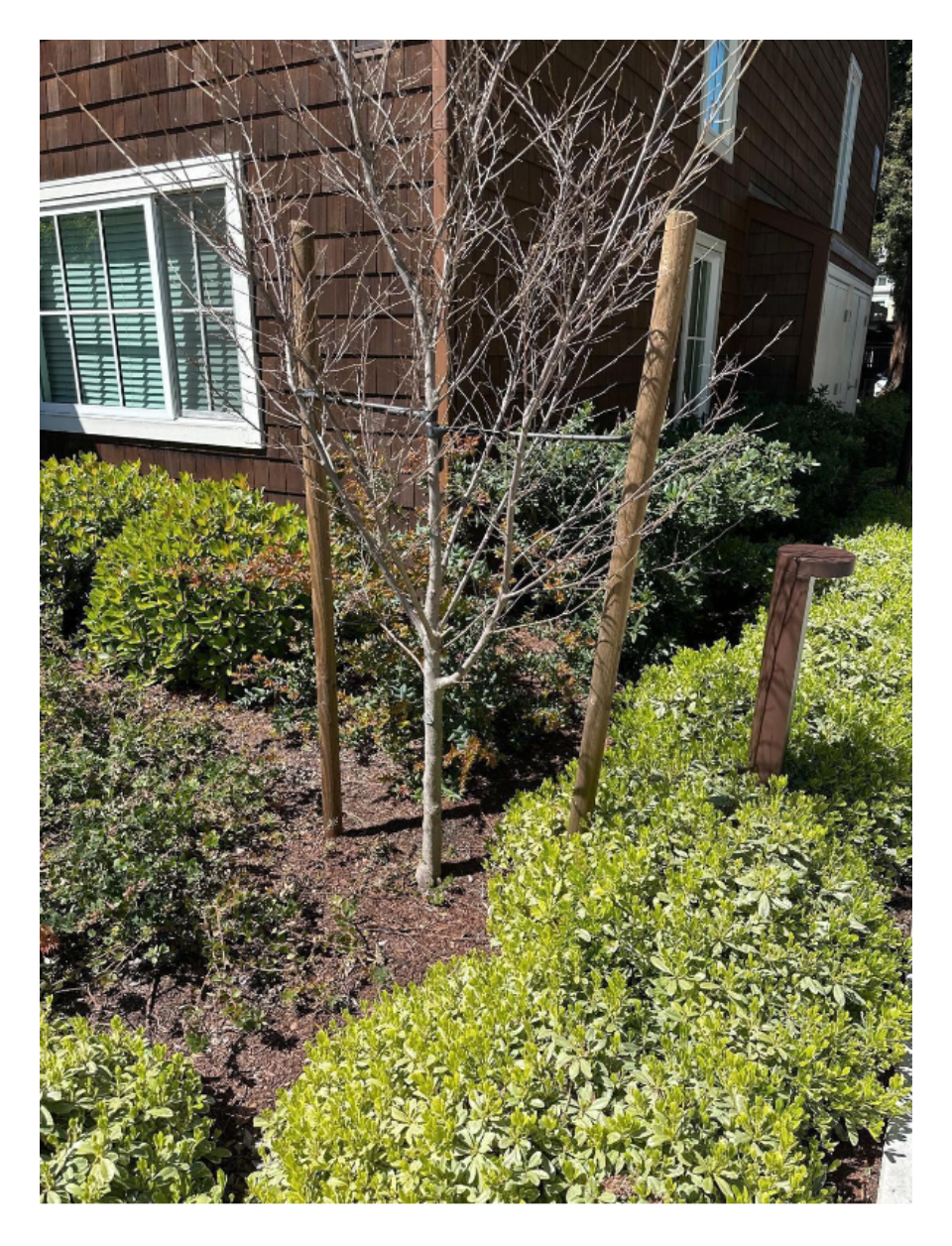}}
\vspace{\medskipamount}

\textbf{Question: }How many wooden poles are in this photo?

\noindent\rule{\linewidth}{0.5pt}

\textbf{Ours 4B: }
{\scriptsize\ttfamily

\textbf{\textcolor{blue}{<think>}}

To determine the number of wooden poles in the photo, let's analyze the image step by step:
 
 1. **Foreground Analysis**: The image shows a small tree with a thin trunk. There are two visible wooden poles supporting the tree. These poles are positioned on either side of the tree, providing stability.
 
 2. **Background and Surroundings**: The background includes a building with windows and some greenery. There are no additional wooden poles visible in the background or elsewhere in the image.
 
 3. **Counting the Poles**: Based on the visible elements, there are exactly two wooden poles supporting the tree.
 
 Given this analysis, the correct answer is:
 
 **C. There are 2 wooden poles.**

\textbf{\textcolor{blue}{</think>}}\\

The final answer is \textbf{\textcolor{tablegreen}{$\boxed{C}$}}.
}

\end{tcolorbox}

\end{document}